\documentclass{article}

\usepackage[preprint]{neurips_2025}

\usepackage[utf8]{inputenc} %
\usepackage[T1]{fontenc}    %
\usepackage{hyperref}       %
\usepackage{url}            %
\usepackage{booktabs}       %
\usepackage{amsfonts}       %
\usepackage{nicefrac}       %
\usepackage{microtype}      %
\usepackage{xcolor}         %

\usepackage{colortbl, xcolor}

\usepackage{subfigure}

\usepackage{wrapfig}

\usepackage{bm}
\usepackage{enumitem}

\usepackage{mathrsfs}
\usepackage{amsmath}
\usepackage{caption}
\usepackage{capt-of}
\usepackage{lipsum}
\usepackage{multibib}
\usepackage{adjustbox}
\newcites{app}{References}
\usepackage{nicefrac}

\usepackage{pgfplots}
\pgfplotsset{compat=1.18}
\usepackage{subcaption}

\usepackage{booktabs}
\definecolor{c1}{HTML}{765D97} 
\definecolor{c2}{HTML}{900C3F}
\definecolor{c3}{HTML}{fc6160}
\definecolor{myblue}{HTML}{E6F3FC} 
\definecolor{mygray}{HTML}{DBE2E9}

\usepackage{multirow}

\usepackage{framed}

\usepackage{amsthm}

\newcommand{\op}[1]{\operatorname{#1}}
\newcommand{\method}[1]{$\operatorname{#1}$}

\definecolor{dark2orange}{rgb}{0.9, 0.4, 0.}
\definecolor{dark2purple}{rgb}{0.4, 0.4, 0.8}

\usepackage{mathtools}

\usepackage{xspace}
\newcommand{\model}{ARMD\xspace}

\newcommand{\modelLong}{Auto-Regressive Masked Diffusion\xspace}

\renewcommand{\log}{\op{log}}

\usepackage{amsmath,amsfonts,bm}

\def\Figref#1{Figure~\ref{#1}}

\def\eqref#1{equation~\ref{#1}}

\def\1{\bm{1}}

\def\veta{{\boldsymbol{\eta}}}
\def\vphi{{\boldsymbol{\phi}}}

\def\vg{{\boldsymbol{g}}}

\def\vk{{\boldsymbol{k}}}

\def\vm{{\boldsymbol{m}}}

\def\vq{{\boldsymbol{q}}}

\def\vv{{\boldsymbol{v}}}

\def\vx{{\boldsymbol{x}}}
\def\vy{{\boldsymbol{y}}}
\def\vz{{\boldsymbol{z}}}

\def\mG{{\bm{G}}}
\def\mH{{\bm{H}}}

\def\mK{{\bm{K}}}

\def\mM{{\bm{M}}}

\def\mO{{\bm{O}}}

\def\mQ{{\bm{Q}}}

\def\mV{{\bm{V}}}
\def\mW{{\bm{W}}}
\def\mX{{\bm{X}}}
\def\mY{{\bm{Y}}}

\DeclareMathAlphabet{\mathsfit}{\encodingdefault}{\sfdefault}{m}{sl}
\SetMathAlphabet{\mathsfit}{bold}{\encodingdefault}{\sfdefault}{bx}{n}

\input{Definitions.tex} %

\title{Auto-Regressive Masked Diffusion Models}

\author{
    Mahdi Karami\thanks{ 
    Correspondence to \texttt{mahdikaramia@gmail.com}. This work was conducted outside Google. \\
    Accepted to the 29\textsuperscript{th} International Conference on Artificial Intelligence and Statistics (AISTATS) 2026.
    } \\
  School of Computer Science\\
  University of Waterloo, ON, Canada\\
    \& Google Research\\
  \And 
    Ali Ghodsi\\
  School of Computer Science\\
  University of Waterloo, ON, Canada\\
}

\begin{document}
\maketitle
\begin{abstract}
Masked diffusion models (MDMs) have emerged as a promising approach for language modeling, yet they face a performance gap compared to autoregressive models (ARMs) and require more training iterations.
In this work, we present the \modelLong (\model) model, an architecture designed to close this gap by unifying  the training efficiency of autoregressive models with the parallel generation capabilities of diffusion-based models.
Our key insight is to reframe the masked diffusion process as a block-wise causal model. This perspective allows us to design a strictly causal, permutation-equivariant architecture that computes all conditional probabilities across multiple denoising steps in a single, parallel forward pass.
The resulting architecture supports efficient, autoregressive-style decoding and a progressive permutation training scheme, allowing the model to learn both canonical left-to-right and random token orderings.
Leveraging this flexibility, we introduce a novel \textit{strided parallel generation} strategy that accelerates inference by generating tokens in parallel streams while maintaining global coherence.
Empirical results demonstrate that \model achieves state-of-the-art performance on standard language modeling benchmarks, outperforming established diffusion baselines while requiring significantly fewer training steps. 
Furthermore, it establishes a new benchmark for parallel text generation, effectively bridging the performance gap between parallel and sequential decoding.
\end{abstract}

\section{Introduction} \label{sec:intro}

Autoregressive Models (ARMs) have demonstrated significant success across various sequence modeling tasks, particularly in language modeling with prominent models like GPT~\citep{openai2023gpt4} and LLaMA~\citep{touvron2023llama}. 
The foundational principle of ARMs lies in factorizing the joint probability distribution of a sequence into a product of conditional probabilities using the chain rule~\citep{larochelle2011neural, bengio2000taking}, typically following a fixed, left-to-right generation order. This  enables efficient parallel training and has achieved state-of-the-art performance in diverse domains beyond text, such as image generation~\citep{van2016pixel}, audio synthesis~\citep{van2016wavenet}, and graph modeling~\citep{liao2019efficient,karami2024higen}.
Despite their success, ARMs suffer from the inherent sequential nature of their generation process, leading to slow sampling, especially for long sequences, and challenges for tasks requiring bidirectional context or non-sequential reasoning~\citep{berglund2023reversal, ding2023causallm, bachmann2024pitfalls}.

In parallel, diffusion models have emerged as a powerful generative modeling alternative,  achieving remarkable success in continuous domains such as image generation~\citep{sohl2015deep, ho2020denoising, song2019generative}. 
A key advantage of these models is their ability to perform parallel sampling during inference, which, combined with their high degree of controllability, makes them well-suited for conditional and guided generation tasks. 
This potential has motivated a renewed focus on extending diffusion principles to discrete data modalities~\citep{austin2021structured, lou2023discrete}, including text~\citep{gulrajani2023likelihood,ou2024your,lou2023discrete,nie2025large} and graph~\citep{vignac2022digress} and biological sequences~\citep{sahoo2024simple}.

Despite these advancements, discrete diffusion models have struggled to match the performance benchmarks set by ARMs in language modeling. 
Moreover, they typically require longer training schedules and a large number of denoising steps during inference, which can severely hinder their sampling efficiency~\citep{austin2021structured, lou2023discrete}. 
Addressing these limitations has become an active area of research, with recent efforts aiming to improve both the scalability and expressiveness of discrete diffusion models.

\paragraph{Contributions.}
To address these challenges, we introduce the  \textit{\modelLong ({\model})} model, which unifies the strengths of diffusion-based learning with the efficiency of autoregressive models.
First, we  reframe masked diffusion models (MDMs) as block-wise causal models. This perspective enables the parallel evaluation of all conditional probabilities within a sequence using a single network call, significantly improving training efficiency.
Leveraging this insight, we propose a causal, permutation-equivariant, attention-based architecture that generalizes traditional autoregressive models. 
This flexible design supports hybrid training, allowing the model to learn effectively from both canonical left-to-right and random orderings. %
\model is also compatible with key-value caching, enabling efficient, autoregressive-style decoding during inference.  
Furthermore, we introduce a novel \textit{strided parallel generation} strategy that accelerates sampling by generating tokens in parallel streams and bridges the performance gap with sequential generation.
The resulting model achieves state-of-the-art performance across standard language modeling benchmarks, while requiring significantly fewer training steps than competing diffusion-based methods.

\section{Background} \label{sec:background}

\begin{figure}[t]
\begin{minipage}[t]{0.63\textwidth}
    \centering
    \begin{minipage}[t]{0.99\textwidth}
    \centering
        \includegraphics[trim= 200 25 200 25,clip, angle=90,origin=c, width=0.9\linewidth]{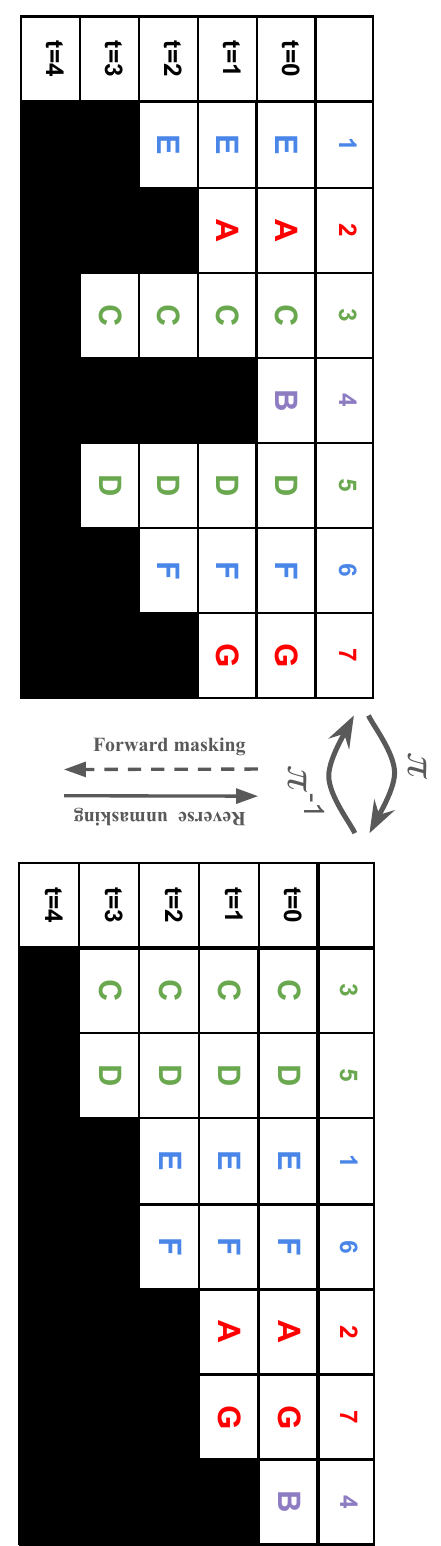}
    \end{minipage}%
    \\
    \vskip -75pt
    \begin{minipage}[t]{0.49\textwidth}
        \centering
        \includegraphics[width=0.9\linewidth]{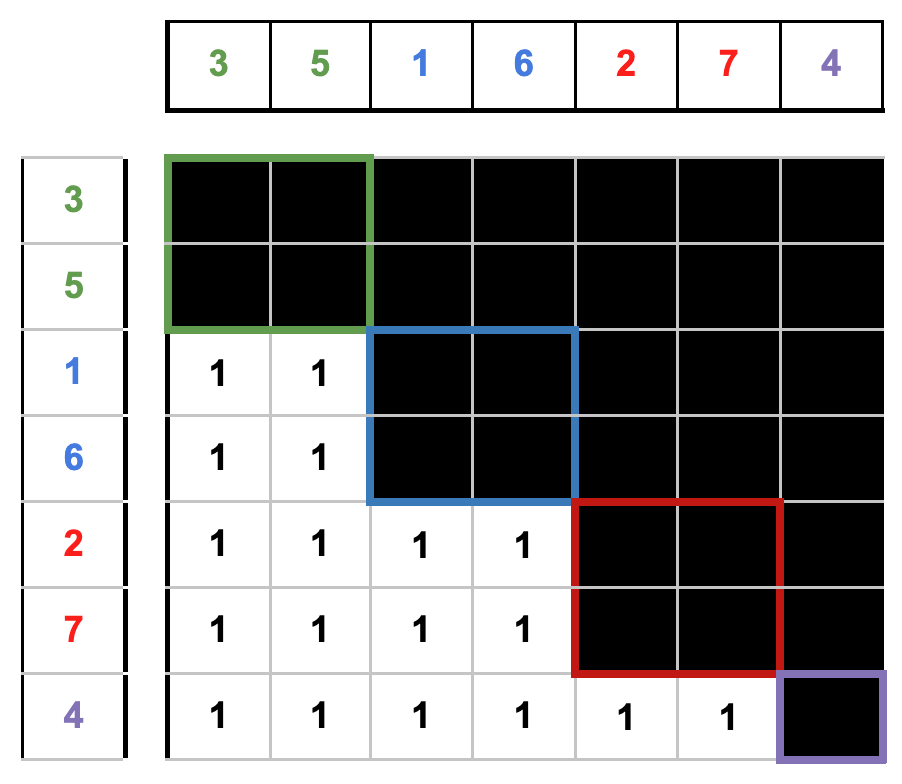}
    \end{minipage}%
    \begin{minipage}[t]{0.49\textwidth}
        \centering
        \includegraphics[width=0.9\linewidth]{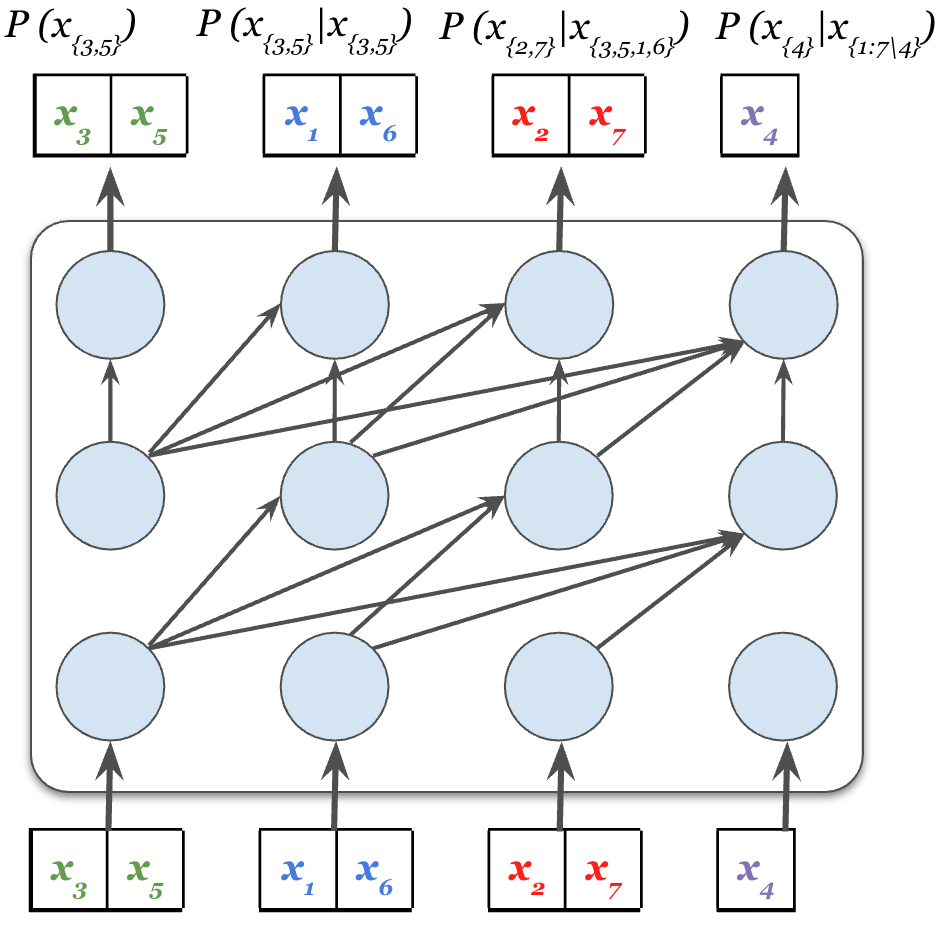}
    \end{minipage}%
    \captionof{figure}{ \footnotesize \label{fig:masking}
    \textit{(Top Left)} An instance of the masked diffusion process on $\vz_{1:N}$ for $T=4$ steps.
    \textit{(Top Right)} The causal patterns derived by permuting the sequence $\vx_{1:N} = \pi(\vz_{1:N})$. The input sequence is partitioned into blocks according to the reverse masking order: $\mathcal{X}(1)=\{3, 4\}$, $\mathcal{X}(2)=\{1, 6\}$, $\mathcal{X}(3)=\{2, 7\}$, and $\mathcal{X}(4)=\{4\}$.
    \textit{(Bottom Left)} The resulting strictly causal attention mask.
    \textit{(Bottom Right)} A model instance composed of a single strictly causal layer followed by two causal layers, forming a deep strictly causal architecture.
    }
\end{minipage}
    \hfill
\begin{minipage}[t]{0.35\textwidth}
    \setlength{\tabcolsep}{1.1pt} %
    \renewcommand{\arraystretch}{2.}
    \centering
    \vskip -55pt
    \resizebox{1.02\linewidth}{!}{
    \begin{tabular}{llccccccc}
    \footnotesize
        1.~ & \texttt{The} & \_ & \_ & \_ & \_ & \_ & \_ & \_ \\
        2. & \texttt{The} & \_ & \_ & \_ & \texttt{jumps} & \_ & \_ & \_ \\
        3. & \texttt{The} & {\color{blue} \texttt{quick}} & \_ & \_ & \texttt{jumps} & {\color{blue} \texttt{over}} & \_ & \_ \\
        4. & \texttt{The} & {\color{blue} \texttt{quick}} & {\color{red} \texttt{brown}} & \_ & \texttt{jumps} & {\color{blue} \texttt{over}} & {\color{red} \texttt{the}} & \_ \\
        5. & \texttt{The} & {\color{blue} \texttt{quick}} & {\color{red} \texttt{brown}} & {\color{green!60!black} \texttt{fox}} & \texttt{jumps} & {\color{blue} \texttt{over}} & {\color{red} \texttt{the}} & {\color{green!60!black} \texttt{dog}} \\
    \end{tabular}
    }
    \vskip 10pt
    \captionof{figure}{ \footnotesize
    Visualization of \textit{strided parallel generation}. Steps 1--2 represent the sequential generation of the stream heads. Steps 3--5 illustrate the parallel generation of subsequent tokens, where tokens of the same color are generated simultaneously. \autoref{fig:masking_strided_gen} depicts the corresponding diffusion process and attention mask.
    }
    \label{fig:gen_steps}
\end{minipage}
\end{figure}

\paragraph{Diffusion Generative Models.}
Diffusion models represent a class of probabilistic generative models that learn a data distribution by systematically reversing a process that corrupts data with noise.  
These models consist of two core components: a \textit{forward diffusion process} that progressively adds noise to a clean data sample $\vz^0 \sim p_{data}(\vz^0)$, and a reverse \textit{denoising process}, 
which is trained to invert this corruption. 
In the continuous domain, the forward process typically adds Gaussian noise over a chain of $T$ timesteps, producing a series of increasingly noisy latent variables
$[\vz^1, \vz^2, \dots, \vz^T]$, where $\vz^T$ approximates a simple prior distribution, such as an isotropic Gaussian.
The forward process is defined as a Markov chain specified by a predefined  conditional probability 
$q(\vz^t \mid \vz^{t-1})$. Consequently, the  joint distribution of the latent variables can be factorized as:
$
q(\vz^1, \vz^2, \dots, \vz^T \mid \vz^0) = \prod_{t=1}^T q(\vz^t \mid \vz^{t-1}),
$
For sufficiently large $T$, $q(\vz^T)$ gets close to a noise prior distribution.
The backward denoising process, parameterized by a neural network $p_\theta(\vz^{t-1} \mid \vz^t, t)$, is trained to approximate the time-reversed dynamics of this diffusion chain and generate clean samples from the data distribution.

In \textit{discrete domains}, such as text,  where data points belong to a finite set of $K$ categories, each sample is represented by a one-hot vector $\vz^0 \in \{0,1\}^K$. %
Following the framework proposed by~\citet{austin2021structured}, the discrete forward  process can be modeled using a sequence of stochastic transition matrices $\{\mQ^t\}_{t=1}^T$, where $[Q^t]_{ij}$ specifies the probability of transitioning from state $j$ (the $j$-th class) to state $i$ (the $i$-th class).
The conditional distribution at time step $t$ is thus defined by a categorical distribution:
$
q(\vz^t \mid \vz^{t-1}) =  \op{Cat}(\vz^t;~ \mQ^t \vz^{t-1}).
$

A common parameterization for the transition matrix interpolates between the data and a stationary noise distribution $\vphi \in \Delta^{K-1}$ (the $K$-simplex):
$
\mQ^t = \beta^t \eye + (1 - \beta^t) \vphi \one^\top,
$
where $\beta^t \in [0,1]$ controls the noise level at step $t$ and $\vphi$ is typically a uniform distribution with $\one$ denoting an all-one vector.
This equation leads to a closed-form expression for the transition:
\begin{align*}
q(\vz^t \mid \vz^{t-1}) &= \op{Cat}(\vz^t;~ \beta^t \vz^{t-1} + (1 - \beta^t)\vphi), \quad
\nonumber\\ 
q(\vz^t \mid \vz^{0}) &= \op{Cat}(\vz^t;~ {\alpha}^t \vz^{0} + (1 - {\alpha}^t) \vphi), 
\end{align*}
where ${\alpha}^t = \prod_{i=1}^t \beta^i$ is the cumulative probability that a token remains in its initial state after $t$ steps. This interpolation between the clean data and a fixed noise distribution is analogous to additive Gaussian noise in the continuous setting~\citep{sahoo2024simple}.

\paragraph{Masked Diffusion Models.}
A simple yet effective variant of discrete diffusion, particularly relevant for sequence data like text, is the Masked Diffusion Model (MDM). This approach introduces an additional special state, denoted by \verb|[mask]|, therefore $\vz_t \in \{0,1\}^{K+1}$, with the $(K+1)$-th dimension corresponding to the mask token.
The noise process is designed such that the \verb|[mask]| state is absorbing: 
once a token is masked, it remains masked in all subsequent steps,
and at step $T$, all tokens are fully masked.
In this setting, the noise distribution $\vphi$
is replaced by a one-hot vector $\vm$ with a $1$ at index $K+1$. Thus, the forward transition simplifies to:
$q(\vz^t \mid \vz^{0}) = \op{Cat}(\vz^t;~ {\alpha}^t \vz^{0} + (1 - {\alpha}^t) \vm) $.
MDMs have often achieved state-of-the-art performance over a range of discrete generative tasks~\citep{austin2021structured, lou2023discrete}.

The reverse process, $p_\theta(\vz^{t-1}_j | \vz^t_{1:N}) $,  is trained to iteratively recover the clean sequence $\vz^0_{1:N}$ from a fully masked input.  This is achieved by optimizing the Evidence Lower Bound (ELBO) on the log-likelihood of the data~\citep{austin2021structured}:
\begin{align*}
    & -\log p_{\theta}(\vz^0)  \le
    \mathcal{L}_{\text{prior}} + \mathcal{L}_{0} 
+ \underbrace{
\sum_{t=1}^{T} 
\mathbb{E}_{ q} 
\left[ D_{\text{KL}}\left( q(\vz^{t-1} | \vz^0, \vz^t) \, \| \, p_\theta(\vz^{t-1} \mid \vz^t_{1:N} ) \right) \right]
}_{\mathcal{L}_{\text{diff}}} 
\end{align*}
where $D_{\text{KL}}$ is the Kullback–Leibler divergence,
$
\mathcal{L}_{0} = \mathbb{E}_{q(\vz^0)} 
\left[ -\log p_\theta(\vz^0 | \vz^1) \right]
$ is the reconstruction loss for the final denoising step
and
$
\mathcal{L}_{\text{prior}} = D_{\text{KL}}\left(q(\vz^T \mid \vz^0) \,\|\, p(\vz^T) \right) 
$ measures the discrepancy between the forward process's final distribution and the noise prior. 
For MDMs where the forward process deterministically leads to a fully masked state, $q(\vz^T \mid \vz^0) = p(\vz^T) =\vm $, then $\mathcal{L}_{\text{prior}}=0$, and the discrete time loss simplifies to~\citep{sahoo2024simple}:  
\begin{align} \label{eq:VLoss_diff}
\mathcal{L}_{\text{diff}} =
\sum_{t=1}^{T} \mathbb{E}_{q} 
\Bigg[ 
\frac{\alpha^t -\alpha^{t-1}}{1-\alpha^t}
\underbrace{
\sum_{\{ j \mid \vz^t_j = \vm \} } \log 
p_\theta(\vz^{t-1}_j \mid \vz^t_{1:N}) 
}_{{L}_t}
\Bigg]
\end{align}
Here, the inner loss term  ${L}_t$ is a cross entropy objective computed specifically on the subset of positions that were just masked at step $t$, thereby focusing the prediction only on unmasking them.

\textit{Autoregressive Models (ARMs)}, on the other hand, factorize the joint probability of a sequence into a product of univariate conditionals using the chain rule. Given a predefined ordering of the variables (e.g., left-to-right), the log-likelihood is decomposed as:
$$
\log p(\vx_{1:N}) = \sum_{t=1}^{N} \log p(\vx_t \mid \vx_{<t}),
$$
where each token $\vx_t$ is predicted based on the preceding tokens in the sequence ($x_{<t}$). 
Thanks to causal masking and teacher forcing, ARMs can evaluate the full log-likelihood for a sequence in a single forward pass, allowing for highly parallelized and efficient training. 
In contrast, the training objective for MDMs \eq{eq:VLoss_diff} requires  sampling a different timestep $t$ for each training example and making a prediction with the model. 
To optimize the full objective, this requires many neural network calls, each evaluating  a portion of likelihood corresponding to a denoising step, ${L}_t$, making its training less efficient per epoch.

However, the MDM training process can be reformulated to parallelize its objective. The key insight is that the random masking schedule induces a causal structure by reordering the sequence.
Formally, for a sample from masking process, let $\tau(j) \in \{1, ..., T\}$ denote the \textit{masking timestep} at which the $j$-th token of the original sequence $\vz_{1:N}$ is changed to \verb|[mask]|. 
We define a permutation $\pi$ over the token indices that sorts the tokens 
based on  their masking times in descending order:
$\tau(\pi(1)) \ge \tau(\pi(2)) \ge ...\ge \tau(\pi(N))$.
This reorders the sequence so that tokens masked last appear first in the new ordering. 
Let the resulting permuted sequence be denoted by 
$\vx_{1:N} = \pi(\vz_{1:N}) = ( \vz_{\pi(1)}, \vz_{\pi(2)}, ..., \vz_{\pi(N)})$.
This reordering induces a partitioning of $\vx_{1:N}$  into $T$ non-overlapping (and possibly variable-length) blocks:
$\mX = [\mathcal{X}^{(1)}, \mathcal{X}^{(2)}, \ldots, \mathcal{X}^{(T)}]$, sorted by their generative (unmasking) order. 
Each block $\mathcal{X}^{(b)} \in \mathbb{R}^{n_b \times d}$ contains $n_b$ tokens, and $\sum_{b=1}^{T} n_b = N$. 
The block index of token $n$, denoted by $\mathcal{B}(n) \in \{1, \dots, T\}$, is then $\mathcal{B}(n) := T-\tau(n)+1$.
As illustrated in \Figref{fig:masking}, this reordering naturally reveals \textit{a block-wise causal} (block triangular) structure: predicting the tokens in a block $\mathcal{X}^{(t)}$ is conditionally dependent only on tokens in preceding blocks $\mathcal{X}^{(<t)}$.
This allows the diffusion loss to be represented in an autoregressive style:
\begin{align} \label{eq:loss_diff_AR}
\mathcal{L}_{\text{diff}} = \mathbb{E}_{q} 
\Big[ 
\sum_{t=1}^{T-1} \!\! 
\gamma(t)
\!\!\!\! \sum_{ n \in \mathcal{B}(t)  } \!\! \!\! \!\! - \log 
p_\theta(\vx_n \mid [\vx_i \mid  {\mathcal{B}(i) \le T-t} ]) 
\Big]
\end{align}
where $\gamma(t)$ is the reweighting factor that reflects the diffusion schedule.
This reformulation of the masked diffusion process as a causal generative model reveals a conceptual and structural connection between MDMs and ARMs, enabling parallel evaluation of all conditionals in the loss by a single forward pass. 
Importantly, the conditional probabilities in the diffusion objective \eq{eq:VLoss_diff} are invariant to the permutation of the evidence set. Additionally, ${L}_t$ is invariant to the ordering of tokens within the target block $\mathcal{B}(t)$. Therefore, this structure necessitates a \textit{permutation-equivariant} sequence model to parameterize these conditionals.   

In the following section, we introduce a deep, permutation-equivariant, and strictly causal sequence model tailored for efficient training of masked diffusion models.
Unlike MLM or BERT models that represent masking by inserting explicit \verb|[mask]| tokens into the input, the proposed architecture controls information flow via a carefully designed attention mask. 
This mechanism restricts output representations from accessing input positions designated as masked, directly enforcing the conditional dependencies required by the denoising process.

\section{Strictly Causal Models} \label{sec:Method}

As explained in the previous section and in \eqref{eq:loss_diff_AR}, the conditional probability is parameterized by a sequence model, where the output at any given position $n$ depends exclusively on preceding input blocks—\textit{i.e.} the present and all future blocks are masked so they cannot influence the current output. We call any network satisfying this property \textit{strictly causal} which is illustrated in \Figref{fig:masking}.
Formally, a \emph{strictly causal layer} is defined as $y_n \;=\; f^{sc}(x_{<\mathcal{B}(n)})$.
This is distinct from a standard \emph{causal layer}, where the output can also depend on the current input block, i.e. 
$ y_n \;=\; f^{c}(x_{\le \mathcal{B}(n)})\,. $

Strictly causal systems form a proper subclass of causal systems which are also physically realizable. Importantly, composing a strictly causal layer with any number of causal ones ({e.g.}, $f^{c}\circ f^{sc}$) yields a network that preserves overall strict causality.
This property allows us to construct arbitrarily deep, strictly causal networks by stacking arbitrarily many causal layers and interleaving just one strictly causal layer (block) which is particularly advantageous in designing our deep neural network architectures.\footnote{It's important to note that within our framework, causality is defined by the processing order of information (or absorbing noise), not necessarily the temporal order of tokens in the original sequence.
}

\paragraph{Causal Self-Attention.}
To implement this principle in a modern transformer architecture, we adapt the standard self-attention mechanism.
The output of a standard self-attention layer for an input sequence $\mX = [\vx_1, \dots, \vx_N]^\top ~\in \RR^{N \times d}$ (where $\vx_n \in \RR^{d}$) is given by:
$$\mY = \op{SA}(\mX) = \op{Softmax}(\mQ \, \mK^\top ) \, \mV$$
Here, the query, key, and value matrices are linear projections of the input: $\mQ = \mX \mW^q $, $\mK = \mX \mW^k $, and $\mV = \mX \mW^v$, where $\mW^q, \mW^k, \mW^v \in \RR^{d \times d}$. By design, this operation is permutation-equivariant.
Defining attention scores corresponding to token $n$ as 
$A_{(n , i)} = \op{Softmax}_i( \vq_n^\top  \mK^\top)$, 
the output at time $n$ can be written as a weighted sum over the entire sequence:
$
\vy_n = \op{SA}_n(\mX) =  \sum_{i=1}^{N} A_{(n , i)} \vv_i
$.

To enforce causality at the block level, we partition the input sequence $\mX$ into $T$ non-overlapping blocks, $\mX = [\mathcal{X}^{(1)}, \mathcal{X}^{(2)}, \ldots, \mathcal{X}^{(T)}]$, and introduce a block-wise causal mask, $\mM^{c}$.
This mask modifies information flow and restricts each token to attend only to tokens in the current or preceding blocks:
\begin{align} \label{eq:attn_c}
    \vy_n &= \operatorname{SA}^{c} \left( [\vx_i \mid  {\mathcal{B}(i) \le \mathcal{B}(n)} ] \right) [n] 
    = 
    \sum_{i :~ \mathcal{B}(i) \le \mathcal{B}(n)} \op{Softmax}_i( {\vq_n^\top \mK^\top } + {M}_{(n , i)}^{c}  ) \, \vv_i,
\end{align}
where, the mapping $\mathcal{B}(n) \in \{1, \dots, T\}$ returns the block index of token $n$. The output sequence in matrix form is then:
$$\mY = \op{SA}^{{c}}(\mQ, \mK, \mV ; \mM^{c}) = \op{Softmax}({\mQ \, \mK^\top }+ \mM^{c} ) \, \mV$$
where $\mM^{c} \in \{0, -\infty\}^{N \times N}$ is the causal mask matrix
\footnote{Refer to \autoref{apdx:not_def} for a detailed notation  definition.}.
This results in a block-lower-triangular attention matrix, allowing full attention within each block (intra-block) while  prohibiting attention to any future blocks.

In this block-wise causal attention, the output $\vy_n $ for token $n$ is generated through the interaction of its corresponding query vector $\vq_n$ with the key and value context of its current block and any preceding blocks.
Hence, the output $\vy_n $ depends on the current block, $\mathcal{B}(n)$,  in two ways: 1) through its own query vector $\vq_n$, and 2) by attending to the key-value cache within the same block (\ie $\{\vk_i, \vv_i \mid i \in \mathcal{B}(n) \}$. This allows for a full intra-block interaction.

To construct a \textit{strictly causal attention layer}, we must completely eliminate any dependence on the current block.
This involves two key modifications to prevent any information leakage from the current block. 
First, we apply a strictly causal attention mask, $\mM^{sc}$, that permits attention only to keys/values from preceding blocks. 
Second, we employ a representation for the query that depends only on the past blocks (instead of the standard query $\vq_n$ which is a function of the current input $\vx_n$).  Specifically, we define a modified query as a strictly causal function: 
\begin{align}
   \hat{\vq}_n = f_q^{sc}\left( \left\{ \vq_i \mid \mathcal{B}(i) < \mathcal{B}(n) \right\} \right),
\end{align}
where $f_q^{sc}(\cdot)$ is a permutation-equivariant aggregation function (e.g., mean-pooling or learned transformation) that pools information from the queries of previous blocks to ensure $\hat{\vq}_n$ has no dependence on the current block $\mathcal{B}(i)$.

By incorporating these two constraints, the output of the \textit{strictly causal attention}  layer is defined as:
\begin{align} \label{eq:attn_sc}
    \vy_n & = \operatorname{SA}^{sc}\left( [\vx_i \mid  {\mathcal{B}(i) < \mathcal{B}(n)} ]\right) [n] 
    = 
    \sum_{i : \mathcal{B}(i) < \mathcal{B}(n)} \op{Softmax}_i \left(  {\hat{\vq}_n^\top  \mK^\top} + {M}_{(n , i)}^{sc}  \right) \, \vv_i , 
\end{align}
This formulation ensures that both the query and the context it attends to
are derived strictly from inputs preceding the current block, thereby satisfying the condition for strict block-level causality.
In the following, we explore different design choices for constructing the strictly causal query aggregation function,  $f_q^{sc}(\cdot)$. 

\textbf{Shift:} A straightforward approach is to define the modified query at position $n$ using the token immediately preceding it, resulting in a simple one-step shift function:
$\hat{\vq}_n = f_q^{sc}\left( \vq_{n-1} \right)$. 
This operation effectively conditions the query on the most recent token from the past. 
While simple, it is inherently not permutation-equivariant and is only suitable for a fixed forward (left-to-right) processing order. 
Consequently, a deep Transformer architecture that incorporates a single attention layer with this shifted query—while other layers use standard causal attention—collapses into a conventional autoregressive language model, which learns to sequentially generate the next token in a left-to-right order.

To derive a strictly causal and permutation-equivariant  query representation suitable for our problem, we explore alternative architectural designs. Our goal is to minimize extra parameters and computational overhead while maintaining high expressivity. Our strategy involves the following two main components:

\paragraph{Prefix Aggregation:}
First, our approach involves aggregating information from all past tokens (the prefix) into a summary vector that can be used to form the strictly causal query. We introduce a specialized feed-forward layer that computes an intermediate representation, $\vg^0_n$, as a position-aware weighted sum of all inputs from preceding blocks:
\begin{align*} %
\vg^0_n = \operatorname{FF}^{sc}\left( [\vx_i \mid \mathcal{B}(i) < \mathcal{B}(n)] \right)[n] = \! \!\! \!\! 
\sum_{i : \mathcal{B}(i) < \mathcal{B}(n)} H_{(n,i)} \, \vx_i
\end{align*}
where $H_{(n,i)}$  is the weight coupling an input token $i$ to an output token $n$. 
Directly parameterizing the weight matrix $\mH$ is computationally expensive, requiring a quadratic number of parameters and operations relative to the sequence length.
To circumvent this, we employ an efficient parameterization where the weight $H_{(n,i)}$ is defined as a function of the actual positions in the sequence ($\hat{\pi}(i)$ and $\hat{\pi}(n)$). Specifically, we parameterize $H_{(n,i)}$ using the dot-product similarity of their corresponding positional embedding vectors:
$$
H_{(n,i)} = \langle \mathbf{PE}_{\hat{\pi}(n)}, \mathbf{PE}_{\hat{\pi}(i)} \rangle
= \langle \hat{\mathbf{PE}}_{n}, \hat{\mathbf{PE}}_{i} \rangle.
$$
Here, $\mathbf{PE}_{j}$ represents the $j$-th row of the positional embedding matrix $\mathbf{PE} \in \mathbb{R}^{N \times d_{pe}}$, $\hat{\pi}(n)$ is the position index of $n$th token, and $\hat{\mathbf{PE}} = \pi \left(  \mathbf{PE} \right)$ denotes the positional embedding matrix reordered according to the sequence permutation $\pi$.
This parameterization ensures the layer has a \textit{fixed number of parameters}, independent of sequence length.
Moreover, we can cast the computation of $\vg^0_n$ as a \textit{linear attention} form~\citep{katharopoulos2020transformers}, the full block matrix computation can  be implemented efficiently using a masking matrix to enforce strict causality as:
\begin{align} \label{eq:PA_matrix}
\mG^0 \!= \operatorname{PA}^{sc} ( \mX ) \!=\! \left(  ( \hat{\mathbf{PE}} \times \hat{\mathbf{PE}}^\top) \odot \mM^{sc} \right) \! \mX 
\end{align}
where $\mM^{sc} \in \{0, 1\}^{N \times N}$ is a strictly causal mask.
The output, $\vg^0_n$, subsequently serves as the basis for deriving the strictly causal query $\hat{\vq}_n$.
The resulting layer inherits the permutation-equivariance (with respect to the condition set) of the attention.
Furthermore, this linear attention can be reformulated as a recurrent form~\citep{katharopoulos2020transformers}, offering computational benefits: it enables sub-quadratic parallel training through techniques such as chunkwise processing~\citep{hua2022transformer, kacham2024polysketchformer} or parallel scan algorithms~\citep{blelloch1990prefix, smith2023simplified}, and enjoys a constant-time complexity per generated token at inference.

\paragraph{Two-Stream Strictly Causal-Attention}
To efficiently construct a deep strictly causal architecture—without introducing additional parameters—we adopt a \textit{two-stream attention architecture}.
At each layer $l$, this design maintains and updates two parallel representations: a causal representation $\mX^l$ and a strictly causal representation $\mG^l$.
The key idea is that both streams share the same attention parameters and key-value context, but operate with different masking and query inputs: 
the causal stream allows access to the current and past tokens, while the strictly causal stream  uses the stricter mask and  derives its query $\hat{\mQ}$ from the previous strictly causal output $\mG^{l-1}$. Formally, the update rules for layer $l$ are:
\begin{align}
\mX^{l} &= \op{SA}^{c}_{\theta}(\mX^{l-1}; \mM^{c}) \!=\! \op{Attn}\left(\mQ, \mK, \mV; \mM^{c} \right), \\
\mG^{l} &= \op{SA}^{sc}_{\theta}(\mG^{l-1}, \mX^{l-1}; \mM^{sc}) \!=\! \op{Attn}\left(\hat{\mQ}, \mK, \mV; \mM^{sc} \right) \nonumber 
\end{align}
where the query, key, and value projections are defined as:
$\mQ = \mW^q \mX^{l-1},~ 
\mK = \mW^k \mX^{l-1}, ~
\mV = \mW^v \mX^{l-1}, ~
\hat{\mQ} = \mW^q \mG^{l-1}$.
At the final layer of this two-stream chain, the model outputs the strictly causal representation $\mG^{L^{2s}}$.
By leveraging both causal and strictly causal representations throughout this chain, the model is able to produce a deeply contextualized and expressive strictly causal output.  
A related two-stream mechanism was previously shown effective in XLNet~\citep{yang2019xlnet}, which generalized autoregressive pretraining by unifying both autoencoding-style and autoregressive objectives.

\begin{figure}[t]
\begin{minipage}{0.49\textwidth}
\centering
    \begin{adjustbox}{width=.65\linewidth}
        \includegraphics[]{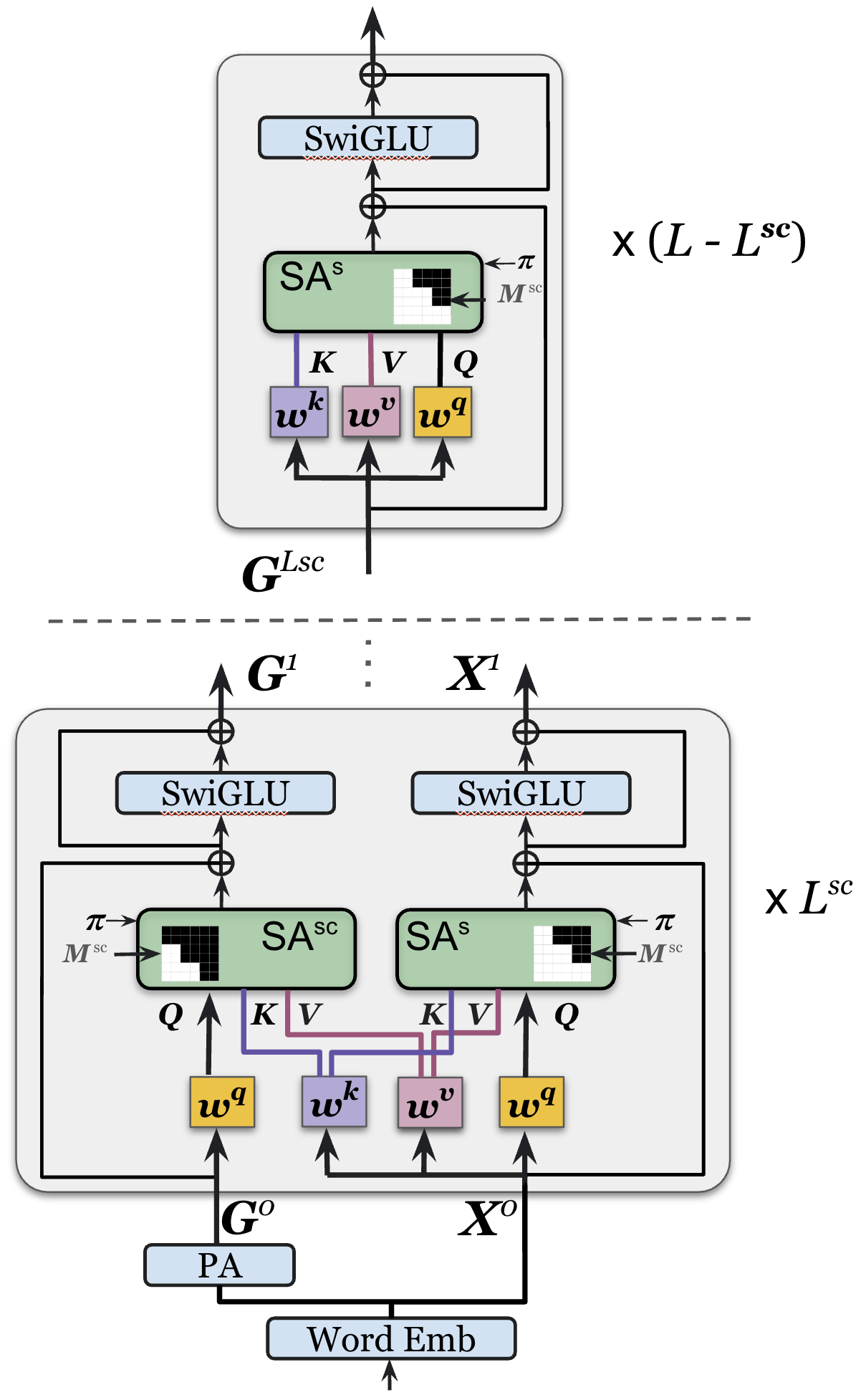}
    \end{adjustbox}
\end{minipage}%
\hfill
\begin{minipage}{0.49\textwidth}
\centering
\captionof{figure}{  \label{fig:deep_arch}
\footnotesize
Schematic of the deep strictly causal architecture. The model is composed of $L^{2s}$ two-stream attention layers (causal and strictly causal streams), followed by $L - L^{2s}$ causal layers. 
The strictly causal stream ($\mG^l$) captures rich contextual and positional representations using only past block information, while the causal stream ($\mX^l$) serves as a shared source of key and value features for the attention layers in both streams. The two streams differ in their query input and masking operations.
The final output is taken from the top layer, enabling the model to parametrize the conditional distribution $p_\theta(\vx_n \mid \{\vx_i \mid \mathcal{B}(i) < \mathcal{B}(n)\})$ while preserving permutation equivariance with respect to the condition set. Layer-Norm, drop-out, and output projections of the attention layers are dropped for simplicity. 
}
\end{minipage}%
\vskip -20pt
\end{figure}

\paragraph{Model Architecture.} 
The final model is composed of two components: (i) an $L^{2s}$-layer two-stream stack and (ii) a subsequent $(L - L^{2s})$-layer standard causal stack.
This composition yields a deep strictly causal model that captures the conditional probability
$
p_{\theta} \left( \vx_n \,\middle|\, [\vx_i \mid \mathcal{B}(i) < \mathcal{B}(n)] \right)
$, ensuring that each token $\vx_n$ is generated exclusively based on preceding blocks. 
Moreover, relative position information is integrated into all attention operations using Rotary Positional Embedding (RoPE)~\citep{su2024roformer}.
As noted in recent works~\citep{ou2024your,zheng2024masked}, it is unnecessary to explicitly parameterize time-dependence in MDMs.
A schematic illustration of the full architecture is shown in \Figref{fig:deep_arch}.

\begin{remark}
As discussed above, this design is inherently based on causal attention mechanisms, thereby enabling efficient key-value (KV) caching for fast decoding—a key advantage over BERT-style models commonly used in MDMs.
Moreover, since the architecture generalizes autoregressive models (ARMs), it naturally supports hybrid training schemes such as \textit{progressive permutation training} (further detailed in \autoref{sec:expriments}), allowing the model to learn effectively from both forward (left-to-right) and randomly permuted orderings.
Finally, the same design principles suggest that the architecture can be readily adapted to finetune pretrained Large Language Models (LLMs) into diffusion models with simple modifications, presenting a promising direction for future research.
\end{remark}

\subsection{Strided Parallel Generation}
To leverage the flexibility of \model{} for efficient generation, we introduce a \textit{strided block-parallel (SBP)} sampling strategy. This method partitions the target sequence of length $N$ into $S$ parallel \textit{streams}, enabling simultaneous token generation while preserving global coherence.
Given a parallelism factor $S$, we first partition the sequence into $S$ equal-sized streams, each of length $N/S$. The indices are then permuted to interleave these streams, grouping tokens that share the same relative position within their respective streams. For example: for $N=8, ~S=2$: streams: 
$\mathcal{S}_1 = [1, {\color{blue} 2}, {\color{red} 3}, {\color{green} 4}]$, $\mathcal{S}_2 = [5, {\color{blue} 6}, {\color{red} 7}, {\color{green} 8}]$ 
$\rightarrow$ permutation $\pi = [1, 5, {\color{blue} 2}, {\color{blue} 6}, {\color{red} 3}, {\color{red} 7}, {\color{green}4, 8}]$. 

The generation process proceeds in two distinct phases: 1) We first generate the initial token (the \textit{head}) of each stream sequentially. In the example above, tokens $(1)$ and $(5)$ are generated autoregressively. This establishes a global context anchor for each stream. 
2) \textit{Parallel Block Generation}: Subsequent tokens are generated in parallel groups of size $S$. Specifically, for each relative position $j > 1$, the model generates the set $\{x^{(1)}_j, x^{(2)}_j, \dots, x^{(S)}_j\}$ simultaneously. 
In the example above, the block $({\color{blue} 2}, {\color{blue} 6})$ is generated in parallel, followed by the block $({\color{red} 3}, {\color{red} 7})$. \autoref{fig:gen_steps} illustrates this parallel generation strategy. 

This strategy maximizes the physical distance between tokens generated in parallel. For instance, tokens $({\color{blue} 2})$ and $({\color{blue} 6})$ are separated by $N/S$ positions in the original sequence. This large separation supports the approximate conditional independence assumption required for valid parallel diffusion sampling, i.e., $p(x_2, x_6 \mid \text{context}) \approx p(x_2 \mid \text{context})p(x_6 \mid \text{context})$, while accelerating inference by a factor of $S$. This strided parallel generation is substantially different from the standard autoregressive block-generation used in \method{BD3-LM}~\citep{arriola2025block}.

\section{Related works}
\paragraph{Discrete Diffusion Models.}
A general framework for diffusion models in discrete space, \method{D3PM}, was introduced by \citet{austin2021structured}. 
This framework is built on a Markov forward process defined by arbitrary transition matrices and optimizes a variational lower bound on the data likelihood. \method{D3PM} was initially applied to language modeling at the character level and, while promising in theory, its early applications struggled to match the empirical performance of autoregressive models.
More recently, \citet{lou2023discrete} advanced the field by adapting score matching to discrete domains using a novel score entropy objective. Their model, \method{SEDD}, demonstrated strong performance on standard language modeling tasks, narrowing the performance gap with ARMs by achieving competitive perplexity and efficient sampling in discrete diffusion for language. 
Among the various discrete diffusion strategies developed within these frameworks, masked diffusion models (MDMs) have demonstrated to be a particularly effective approach~\cite{austin2021structured, lou2023discrete}, which revealed a conceptual connection to Masked Language Models (MLMs) like BERT~\citep{devlin2018bert}.

\paragraph{Masked Diffusion Models.} 
Recent works have rapidly advanced MDMs, enhancing their training efficiency and generative quality to achieve performance competitive with ARMs of a similar size. 
These efforts include exploring simplified objectives and properties of absorbing discrete diffusion models~\citep{shi2024simplified,sahoo2024simple, ou2024your}.
For instance, \citet{sahoo2024simple} developed simplified variational objectives while their resulting method supports semi-autoregressive generation that admits efficient generation of sequences of arbitrary length.
Building on \method{SEDD}, \citet{ou2024your} provided a deeper theoretical understanding of absorbing discrete diffusion.
They proposed a reparameterization of absorbing discrete diffusion models to characterize time-independent conditional probabilities of clean data, demonstrating improvements in sampling efficiency and zero-shot perplexity compared to \method{SEDD}. 
The scaling law properties of MDMs on text data have been demonstrated in \cite{nie2024scaling}, which also proposed an unsupervised classifier-free guidance technique~\citep{ho2022classifier} to further improve inference efficiency on text.
Another promising direction for scaling diffusion models involves adapting existing pre-trained AR models to text diffusion models by gradually annealing the attention mask during fine-tuning~\citep{gong2024scaling}, a technique that leverages the connections between AR and diffusion objectives.
Recently, \citet{arriola2025block} proposed block diffusion, a hybrid approach that combines discrete diffusion within token blocks and ARMs across blocks, which improves inference efficiency and supports flexible-length sequence generation.

\paragraph{Continuous Diffusion in Latent Space.} A separate line of research explores continuous diffusion for discrete data by embedding tokens into continuous latent spaces and applying continuous diffusion on these latent representations~\citep{dieleman2022continuous, li2022diffusion, chen2022analog, gulrajani2023likelihood, han2022ssd, lovelace2023latent, graves2023bayesian}. 
However, the process of decoding from continuous back to discrete space remains challenging and is often prone to information loss. Consequently, these latent diffusion models have generally underperformed ARMs on text generation tasks.

\paragraph{Order Agnostic ARMs.}
Order Agnostic ARMs (OA-ARMs)~\citep{uria2014deep} represent a class of generative models that generalizes traditional ARMs by marginalizing over all possible generation orders, thereby addressing the inherent left-to-right bias of ARMs. This generalization allows models to flexibly condition on arbitrary subsets of observed tokens, thereby supporting flexible non-sequential generation.
Recent works have established the underlying connections between MDMs and OA-ARMs. 
For instance, \citet{hoogeboom2021autoregressive} showed that absorbing discrete diffusion models are equivalent to OA-ARMs under mild assumptions, leading to the proposal of autoregressive diffusion models (ARDMs). 
\citet{ou2024your} further  formally unified the training objectives of absorbing discrete diffusion and OA-ARMs, 
bridging the practical gap between these two models. 
From a practical standpoint, since considering all permutations is computationally infeasible, \citet{shih2022training} improve training efficiency and performance by approximating the marginalization through sampling from a small, carefully chosen subset of orderings.
In practice,  both OA-ARMs and MDMs are trained using a collection of forward passes of masked language models (MLMs)— where a bidirectional model is trained to recover a randomly masked subset of tokens at each call.
In contrast, our proposed model enables the parallel evaluation of the entire  likelihood within a single forward pass.

 \begin{table*}[t!]
  \centering
\footnotesize
\caption{\label{tbl:zero_shot_small_med}
    \footnotesize
    \textbf{Zero-shot language modeling perplexity ($\downarrow$) 
    }:
    \method{D3PM}~\citep{austin2021structured},  \method{PLAID}~\citep{gulrajani2023likelihood} and \method{SEDD-Uniform}~\citep{lou2023discrete} are  discrete diffusion models, while \method{SEDD}~\citep{lou2023discrete} and \method{RADD}~\citep{ou2024your}  are masked language models.
The number of training steps for each model is presented in parenthesis. 
 $^*$Results for baseline diffusion models are taken from the reported upper bounds in ~\cite{lou2023discrete,ou2024your}.
The best result for each dataset is shown in \textbf{bold}, and the second-best is \underline{underlined}.
}
\vspace{-5pt}
\setlength{\tabcolsep}{4pt}
\renewcommand{\arraystretch}{1.2}
  \begin{tabular}{lccccc}
\toprule
\quad  \method{Method} & \method{LAMBADA} & \method{WikiText2} & \method{PTB} & \method{WikiText103} & \method{1BW} \\ 
\midrule
\midrule
\multicolumn{6}{l}{\textsl{small size models}} \\
\quad {\method{GPT-2^*}} & \underline{45.04} & 42.43 & 138.43 & 41.60 & 75.20\\
\arrayrulecolor{black!30}\midrule
\quad {\method{D3PM^*}} & 93.47 & 77.28 & 200.82 & 75.16 & 138.92\\
\quad {\method{PLAID~(600K)^*}} & 57.28 & 51.80 & 142.60 & 50.86 & 91.12\\
\quad {\method{SEDD-Uniform ~(400K)^*}} & 65.40 & 50.27 & 140.12 & 49.60 & 101.37\\ 
\arrayrulecolor{black!30}\midrule
\quad {\method{SEDD ~(400K)^*}}& 50.92 & 41.84& \underline{114.24}& 40.62 & 79.29\\
\quad {\method{RADD ~(400K)^*}} & 51.70& {39.98}& \textbf{107.85}& {37.98}& 72.99\\
\arrayrulecolor{black!30}\midrule
\quad {\method{\model~(180K)}}	&	\textbf{44.66}	&	\underline{36.25}	&	130.31	&	\underline{35.66}	&	\underline{53.18}	\\
\quad {\method{\model~(400K)}}	&	45.35	&	\textbf{35.64}	&	123.43	&	\textbf{35.15}	&	\textbf{52.94}	\\

\arrayrulecolor{black!100}\midrule
\multicolumn{6}{l}{\textsl{medium size models}} \\
\quad {\method{GPT-2^*}} & \textbf{35.66}& 31.80& 123.14& 31.39& {55.72}\\
\arrayrulecolor{black!30}\midrule
\quad {\method{SEDD ~(400K)^*}} & 42.77 & 31.04 & 87.12 & 29.98 & 61.19\\
\quad {\method{RADD ~(400K)^*}} & 44.10& 30.60& \textbf{82.08}& 29.29& 60.32\\
\arrayrulecolor{black!30}\midrule
\quad {\method{\model ~(120K)}}	&	40.62	&	\underline{27.57}	&	105.09	&	\underline{27.09}	&	\underline{45.49}	\\
\quad {\method{\model ~(300K)}}	&	\underline{39.08}	&	\textbf{26.06}	&	97.75	&	\textbf{25.55}	&	\textbf{43.91} \\
\arrayrulecolor{black!100}\bottomrule
  \end{tabular}
\end{table*}

\section{Experiments} \label{sec:expriments}
We evaluate the proposed model, \model, on standard language modeling tasks to demonstrate its effectiveness and training efficiency relative to established generative language models.

\paragraph{Setup.}
Following the experimental setup of \citep{lou2023discrete}, we train two versions of \model with parameter counts similar to GPT-2 small (125M) and GPT-2 medium (345M)~\citep{radford2019language}.
As in~\citep{lou2023discrete,ou2024your}, we train \model on the \method{OpenWebText} dataset~\citep{Gokaslan2019OpenWeb}, a large-scale corpus derived from web pages, widely used for training language models. 
Each training iteration uses a batch of 512 sequences, each 1024 tokens long, amounting to approximately 0.5 million tokens per batch.
For the \model architecture, the first half of the transformer layers are configured as the two-stream module, \textit{i.e.} $L^{2s} = L/2$. Further details are provided in \autoref{apdx:exp_details}.

We compare \model against several established and recent diffusion models of similar size, including: \method{GPT-2}~\citep{radford2019language}, \method{D3PM}~\citep{austin2021structured}, \method{PLAID}~\citep{gulrajani2023likelihood}, \method{SEDD}~\citep{lou2023discrete} and \method{RADD}~\citep{ou2024your}. 
Performance is measured using zero-shot perplexity on a suite of widely-used benchmark datasets: 
LAMBADA~\citep{paperno2016lambada}, Penn Tree Bank (PTB)~\citep{Marcus1993BuildingAL}, WikiText2, WikiText103,~\citep{merity2016pointer},  and One Billion Words (1BW)~\citep{chelba2014billion}.

\paragraph{Progressive Permutation Schedule.}
The flexible permutation-equivariant design of  \model  enables it to learn from orders beyond the canonical left-to-right. 
To leverage this, we introduce a \textit{progressive permutation schedule} during training to 
encourage learning random orderings. This strategy begins by training the model exclusively on forward-ordered sequences for an initial phase ($i_{AR}=9K$ iterations).
Subsequently, we gradually introduce random permutations by progressively increasing the number of shuffled tokens from one up to a maximum of $\rho=32$ by iteration $i_{perm}=48K$ and remain the same over the rest of training.
This curriculum enables the model to become increasingly robust to token permutations while still retaining the left-to-right order as the primary structural prior. Since natural language inherently exhibits a strong left-to-right structure, leveraging this inductive bias as guidance ensures that the model retains strong performance in capturing forward dependencies while learning random permutations.

As shown in \autoref{tbl:zero_shot_small_med}, \model achieves state-of-the-art zero-shot performance across most language modeling benchmarks, for both model sizes. Notably, it outperforms most baselines despite being trained for significantly fewer iterations (180K for the small model and 120K for the medium model). 
This result underscores the superior training efficiency that is a key advantage of our proposed model. We provide extended results, ablation studies, and sample generations in \autoref{apdx:exp_details}.

\begin{table}[thb]
    \centering
    \caption{\textbf{Sample Generation vs. Block Size (Parallelism):} Comparison of \model using sequential generation ($T=1024$) versus parallel generation with $T=512$ and $T=256$, corresponding to block sizes of $S=2$ and $S=4$, respectively.
    Performance is reported as Perplexity($\downarrow$) + (Entropy).
    \model's average sampling time per sequence is 3.51s for $S=1$, 1.78s for $S=2$, and 0.93s for $S=4$ on one H100 GPU.
    \textsuperscript{*}Results for the baselines are adopted from \cite{zheng2024masked}.
    }
    \label{tab:parallel_gen}
\setlength{\tabcolsep}{3.pt}
\renewcommand{\arraystretch}{1.3}
    \begin{tabular}{l c c c}
        \toprule
        \textbf{Model} & \textbf{$T\text{=}1024$} & \textbf{$T\text{=}512$} & \textbf{$T\text{=}256$} \\
        & \small{(Sequential)} & \small{($S\text{=}2$)} & \small{($S\text{=}4$)} \\
        \midrule
        \method{AR~Baseline}\textsuperscript{*} & $\sim$36 (8.1) & - & - \\
        \method{SEDD}\textsuperscript{*} & $\sim$106 (8.1) & $\sim$110 (8.1) & - \\
        \method{MDLM}\textsuperscript{*} & $\sim$103 (8.1) & $\sim$113 (8.1) & - \\
        \midrule
        \method{\model} ($\rho\text{=}32, i_{\text{SBP}}\text{=}10\text{K}$) & 38.7 (8.0) & 43.2 (8.0) & 47.8 (8.1) \\
        \method{\model} ($\rho\text{=}32, i_{\text{SBP}}\text{=}50\text{K}$) & 39.1 (8.0) & 40.1 (8.1) & 43.2 (8.1) \\
        \method{\model} ($\rho\text{=}32, i_{\text{SBP}}\text{=}80\text{K}$) & \textbf{36.5} (7.9) & \textbf{37.8} (8.0) & \textbf{39.4} (8.0) \\
        \midrule
        \method{\model} ($\rho\text{=}16, i_{\text{SBP}}\text{=}10\text{K}$) & 50.0 (8.1) & 54.6 (8.1) & 64.7 (8.2) \\
        \method{\model} ($\rho\text{=}16, i_{\text{SBP}}\text{=}50\text{K}$) & 45.7 (8.0) & 46.1 (8.1) & 50.1 (8.1) \\
        \method{\model} ($\rho\text{=}16, i_{\text{SBP}}\text{=}80\text{K}$) & 39.4 (8.0) & 40.5 (8.1) & 43.7 (8.1) \\
        \bottomrule
    \end{tabular}
\end{table}

\paragraph{Parallel Sample Generation.}
We evaluate the generative perplexity of models in the small parameter setting trained on the \method{OpenWebText} dataset.
For \model, we first train a base model for 400K iterations using the progressive permutation schedule (up to $\rho$). This is followed by a short fine-tuning phase of $i_{\text{SBP}}$ steps (where $i_{\text{SBP}} \in \{10\text{K}, 50\text{K}, 80\text{K}\}$), utilizing the \textit{random strided block permutation} (SBP), 
where the sequence is partitioned into $S$ equal-sized blocks and then permuted to interleave these blocks, grouping tokens that share the same relative position.
During this phase, the block size $S$ is sampled randomly from $\{1, 2, 4\}$ at each step to encourage robust parallel generation.  

As the results in \autoref{tab:parallel_gen} demonstrate, while standard masked diffusion models (\method{SEDD}, \method{MDLM}) struggle with high perplexity during parallel generation, \model (particularly with $\rho=32$) 
efficiently generates high-quality, diverse text (quantified by high mean entropy) with only 10K additional training steps. Notably, the total training budget for \model (410K steps) remains lower than that of the AR baseline (500K steps) and standard MDMs (1M steps). Furthermore, increasing $i_{\text{SBP}}$ significantly improves the parallel generation and effectively bridges its performance gap with sequential generation.

\begin{table}[t!]
\centering
\caption{Test perplexities (ppl; $\downarrow$) on LM1B dataset. 
Baseline models are: \method{BERT-mouth}~\citep{wang2019bertMouth}, \method{D3PM}~\citep{austin2021structured}, \method{DiffusionBert}~\citep{he2022diffusionbert}, \method{Diffusion-LM}~\citep{li2022diffusion},
\method{SEDD}~\citep{lou2023discrete}, \method{MDLM}~\citep{sahoo2024simple} and \method{BD3-LMs}~\citep{arriola2025block}.
\textsuperscript{*}Results for baseline models are taken from \cite{lou2023discrete,arriola2025block}.
The number of training steps is presented in parenthesis.
Best diffusion perplexity is shown in \textbf{bold}. }
\label{tab:ppl_lm1b}
\begin{tabular}{lr}
\toprule
\quad \method{Method} & ppl ($\downarrow$) \\
\midrule
\multicolumn{2}{l}{\textit{Autoregressive}} \\
\quad \method{Transformer-X Base}\textsuperscript{*} {{}} & 23.5 \\
\quad \method{Transformer}\textsuperscript{*} {{}} & 22.83 \\
\arrayrulecolor{black!30}\midrule
\multicolumn{2}{l}{\textit{Diffusion}} \\
\quad \method{D3PM-absorb}\textsuperscript{*} {{}} & 82.34 \\
\quad \method{Diffusion-LM}\textsuperscript{*} 
{{}} & 118.62 \\
\quad \method{DiffusionBert}\textsuperscript{*} {{}} & 63.78 \\
\quad \method{SEDD} (1M)\textsuperscript{*} {{}} & 32.68 \\
\quad \method{MDLM} (1M)\textsuperscript{*} {{}} & 31.78 \\
\quad \method{BD3-LMs} $L^\prime$= 16 (1M)\textsuperscript{*} {{}} & 30.60 \\
\quad \method{BD3-LMs} $L^\prime$= 4 (1M)\textsuperscript{*} {{}} & 28.23 \\
\arrayrulecolor{black!30}\midrule
\quad \method{\model} (300K) {{}} &  {23.64}\\
\quad \method{\model} (1M) {{}} &  \textbf{22.36}\\
\arrayrulecolor{black!100}\bottomrule
\end{tabular}
\end{table}

\paragraph{One Billion Words Benchmark.}
We also train language models on One Billion Words (LM1B) dataset ~\citep{chelba2014billion}.
We follow the training and architecture setting of  \cite{he2022diffusionbert, lou2023discrete, sahoo2024simple}: the model size matches GPT-2 small, and the \texttt{bert-base-uncased} tokenizer is applied.
\model employs two-stream modules in the first half of its layers \ie $L^{2s} \text{ = } L/2 \text{ = } 6$.
For progressive permutation schedule of \model, we begin introducing random permutations at iteration $i_{AR} \text{ = } 100\text{K}$, gradually increasing to a maximum of $\rho \text{ = } 8$ by iteration $i_{perm} \text{ = } 300\text{K}$.

As shown in \autoref{tab:ppl_lm1b}, \model achieves the best performance among all diffusion models with significantly fewer training steps.
Notably, it outperforms MDMs with $3\times$ fewer training steps, reconfirming the superior training efficiency of our proposed architecture.

\section{Conclusion}
In this work, we introduced \model, a novel parallel masked diffusion model,  that bridges the gap between diffusion-based learning and autoregressive modeling for language models. 
By reframing the masked denoising process as a block-wise causal problem, we developed  a strictly causal, permutation-equivariant, attention architecture that enables fully parallelized training. 
This flexible design 
supports a progressive permutation schedule for learning from both fixed and random token orderings.
Importantly, our architecture enables a strided parallel generation strategy that accelerates inference while effectively bridging the performance gap between parallel and sequential generation.
Experimental results demonstrate that \model achieves state-of-the-art performance on standard language modeling benchmarks, outperforming both ARM and MDM baselines with significantly fewer training iterations.
By combining the training efficiency of ARMs with the inherent flexibility of diffusion models, \model stands as a scalable and powerful alternative for masked diffusion models.

\clearpage
\bibliographystyle{unsrtnat}
\bibliography{references}

\clearpage
\appendix
\onecolumn
\section{Notation definition} \label{apdx:not_def}

\begin{table}[th]
\begin{center}
\footnotesize
\def\arraystretch{1.3} \tabcolsep=2pt
\begin{adjustbox}{width=.88\linewidth,}
\begin{tabular}{p{1.6in}p{5.4in}}
\toprule 
\textbf{Notations} & \textbf{Brief definition and interpretation} \\
\midrule     \midrule
$N$ & Sequence length (number of tokens). \\
$d$ & Dimensionality of token embeddings. \\
$T$ & Number of diffusion steps (masking timesteps). \\
$K$ & Number of categories (vocabulary size for discrete tokens). \\
$\vz_j = \vz^0_j$ & Clean sample at position $j$, represented as a one-hot vector $\vz^0_j \in \{0,1\}^K$. 
\\
$\vz^t_j$ & Noisy (partially masked) version of token $j$ at timestep $t$ in the diffusion process. \\
$\vz_{1:N}$ & Original input sequence before masking. \\
$\tau(j)$ & Masking timestep when token $j$ is first masked during the forward diffusion process. \\
$\pi,~ {\pi}(\cdot) $ & $i ={\pi}(j)$, Permutation that sorts token indices by $\tau(j)$ in descending order. Also, as a function it permutes a sequence along the sequence dimension. \\
$\hat{\pi}(\cdot)$ & Inverse permutation: $j = \hat{\pi}(i) = \pi^{-1}(i)$. \\
$\mathcal{U}(\cdot)$ & Uniform distribution \\
$\Pi_N$ &Set of all permutations of the indices $\{1, ..., N\}$  \\  
$\vx_{1:N} = \pi(\vz_{1:N})$ & Reordered sequence where tokens masked last come first. $\vx_{1:N} = ( \vz_{\pi(1)}, \vz_{\pi(2)}, ..., \vz_{\pi(N)})$ \\
$\mX = [\mathcal{X}^{(1)}, \ldots, \mathcal{X}^{(T)}]$ & Partition of $\vx_{1:N}$ into $T$ non-overlapping blocks by unmasking order. \\
$\mathcal{X}^{(t)}$ & Block of tokens generated at step $t$ (with size $n_t \times d$), where $n_t$ is number of tokens in $\mathcal{X}^{(t)}$ with $\sum_{t=1}^T n_t = N$. \\
$\mathcal{B}(n)$ & Block index that token $\vx_n$ belongs to. $\mathcal{B}(n) = T - \tau(n) + 1$. \\
$\gamma(t)$ & Reweighting factor for loss at step $t$ (from diffusion schedule). \\
$\one$ & All-ones vector. \\
$\Delta^{K-1}$ & The $K$-simplex: set of all $K$-dimensional probability vectors. \\
$\vphi \in \Delta^{K-1}$  & Limit (stationary) distribution of the diffusion process
\\
$\vm \in \{0,1\}^{K+1}$ & One-hot representation over $K$ categories + 1 special mask token.
\\
$\op{Cat}(\vz;~ \veta)
$ & Categorical distribution over $\vz$ with probabilities of the classes $\veta$. 
\\
$q(\vz^t \mid \vz^{t-1})$ & Forward diffusion (masking) transition probability.
\\
$p_\theta (~ \mid ~)$ & Learned reverse (unmasking) model parameterized by $\theta$. \\
$D_{\text{KL}}(q \, \| \, p)$ & Kullback–Leibler divergence between distributions $q$ and $p$. 
\\
$\vx_{<n}$ & Subsequence of tokens before token $n$ in a predefined order. 
\\
$f^{c}()$ & causal function: $ y_n \;=\; f^{c}(x_{\le n}) $
\\
$f^{sc}()$ & strictly causal function: $ y_n \;=\; f^{sc}(x_{< n}) $
\\
$\mX$ & Input matrix composed of token embeddings: $ \mX= [\vx_1, \dots, \vx_N]^\top ~ \in  \RR^{N \times d}$
\\
$\mQ, \mK, \mV$ & Query, key, and value matrices from linear projections of $\mX$. \\
$\hat{\vq}_n$ & Strictly causal query at position $n$, independent of block $\mathcal{B}(n)$ \\
$\mW^q, \mW^k, \mW^v$ & Linear projection matrices for computing $\mQ, \mK, \mV$. \\
$\mX^l$ & Causal hidden representation at layer $l$ \\
$\mG^l$ & Strictly causal hidden representation at layer $l$ \\
$\op{SA}^{c}(\mQ, \mK, \mV ; \mM^{c})$  & Causal self-attention layer:  $\op{Softmax}({\mQ \, \mK^\top }+ \mM^{c} ) \, \mV. $ \\
$\op{SA}^{sc}(\hat{\mQ}, \mK, \mV ; \mM^{sc})$ & Strictly causal self-attention layer: 
$\op{Softmax}({\hat{\mQ} \, \mK^\top }+ \mM^{sc} ) \, \mV. $ 
\\
$\mM^c, \mM^{sc}$ & Causal and strictly  attention masks: 
$M_{(n , i)}^{sc} \!\! = \!\!
\begin{cases}
0 ~~ \text{if } \mathcal{B}(i) < \mathcal{B}(n) \\
-\infty ~~ \text{otherwise}
\end{cases}\!\!,~ 
M_{(n , i)}^{c} \!\! = \!\!
\begin{cases}
0  ~~\text{if } \mathcal{B}(i) \le \mathcal{B}(n) \\
-\infty ~~\text{otherwise}
\end{cases}
$
\\
$f_q^{sc}$ & Strictly causal function used to compute $\hat{\vq}_n$ \\
$\operatorname{PA}^{sc}(\cdot) = \operatorname{FF}^{sc}(\cdot)$ & Prefix aggregation layer: a strictly causal feedforward layer for aggregating past information. \\
$h_{ni}$ & Weight of $\operatorname{PA}^{sc}(\cdot)$ from position $i$ to position $n$ \\
$\mathbf{PE},~ \mathbf{PE}_{j},~ \hat{\mathbf{PE}}$ & $\mathbf{PE}$: positional embeddings; $\mathbf{PE}_j$: embedding for position $j$; $\hat{\mathbf{PE}}$: permuted embeddings using $\pi$. 
\\
$p_\theta(\vx_n \mid [\vx_i \mid \mathcal{B}(i) < \mathcal{B}(n)])$ & Autoregressive conditional probability of $\vx_n$ given strictly causal context in MDMs. \\
\bottomrule
\end{tabular}
\end{adjustbox}
\end{center}
\end{table}

\section{Extended Discussion} \label{apdx:elaboration}

\subsection{Order Agnostic ARM}
Order Agnostic ARMs (OA-ARMs)~\citep{uria2014deep} compute the data likelihood by marginalizing over all possible generation orders. This approach enables the model to learn conditional generative distributions over any order,  thereby mitigating the left-to-right bias inherent in traditional ARMs, 
and also allowing the model to flexibly condition on arbitrary subsets of observed tokens, supporting non-sequential generation.
The OA-ARM training objective can be reformulated as~\citep{hoogeboom2021autoregressive,shih2022training}:
\begin{align}
\log p(\vz_{1:N}) 
&\ge \mathbb{E}_{\pi \sim \mathcal{U}(\Pi_N)} \sum_{t=1}^{N} \underbrace{\log p(\vz_{\pi(t)} \mid \vz_{\pi(<t)})}_{L_t} \label{eq:AO-ARM_main}\\
&= \mathbb{E}_{\pi \sim \mathcal{U}(\Pi_N)} N \cdot \mathbb{E}_{t \sim \mathcal{U}(\{1, \ldots, N\})} \log p(\vz_{\pi(t)} \mid \vz_{\pi(<t)}) \nonumber \\
&= N \cdot \mathbb{E}_{t \sim \mathcal{U}(\{1, \ldots, N\})} \mathbb{E}_{\pi \sim \mathcal{U}(\Pi_N)} \underbrace{\frac{1}{N - t + 1} \sum_{k \in \pi(\ge t)} \log p(\vz_k \mid \vz_{\pi(<t)})}_{\Tilde{L}_t}
\label{eq:AO-ARM_BERT}
\end{align}
The likelihood upper-bound \eq{eq:AO-ARM_BERT}  highlights the connection between OA-ARMs and masked diffusion models. Recent work by~\citet{ou2024your} formally unifies the training objectives of MDMs and OA-ARMs.
However, training OA-ARMs in \citep{hoogeboom2021autoregressive,shih2022training,ou2024your} involves optimizing a single timestep objective ${\Tilde{L}_t}$ at a time, similar to masked language models like BERT—by masking random subsets of input tokens and training bidirectional models to recover them. Hence they require longer training than ARMs.

In contrast, we adopt the autoregressive-style likelihood \eq{eq:AO-ARM_main} and introduce a strictly causal, permutation-equivariant  model that allows parallel evaluation of all conditionals during training. 
This is achieved through a carefully designed attention masking, described in detail in \S\ref{sec:Method}.

\paragraph{Recent Diffusion Language Models.}
Several recent works have focused on improving the efficiency and scale of Diffusion Language Models (DLMs).
\citet{sahoo2025diffusionDuality} propose a curriculum learning strategy guided by the Gaussian process, and aslo present Discrete Consistency Distillation, which adapts consistency distillation from the continuous to the discrete setting. Together, these techniques significantly enhance both training and sampling efficiency.
\citet{hu2025accelerating} propose a key-value (KV) approximation caching technique and also employs a lightweight, pretrained autoregressive model to guide the token unmasking process. This method improves the efficiency of DLMs by significantly reducing the total number of required denoising iterations.
Recent efforts have focused on scaling DLMs to sizes comparable to autoregressive models. Initial works like DiffuLLaMA~\citep{gong2024scaling}, LLaDA~\citep{nie2025large} successfully scaled DLMs into the billion-parameter range.
The recent Dream model~\citep{dream2025} pushed this frontier further.
It scales up a DLM to 7 billion parameters by using training strategies, such as initialization from a pretrained autoregressive LLM and a context-adaptive, token-level noise rescheduling scheme. The resulting model demonstrates superior planning abilities and inference flexibility.

\paragraph{Advantages of \model and its Limitations.}
As discussed in \S \ref{sec:Method}, the architecture of \model is built on causal attention, enabling efficient decoding and sampling via \textit{key-value (KV) caching}. 
Since it generalizes autoregressive models (ARMs), \model naturally supports hybrid training strategies—such as \textit{progressive permutation training} (\autoref{sec:expriments}) allowing the model to learn from both forward (left-to-right) and random orderings.
Its flexible design also makes it amenable to \textit{fine-tuning pretrained Large Language Models (LLMs) into diffusion models} with minimal architectural changes, opening promising directions for future work. In addition, the model’s permutation-equivariant property and any-order training allow for a range of sampling strategies, as discussed in \S\ref{sec:sample_gen}.

On the other hand, the proposed two-stream attention architecture~\citep{yang2019xlnet}, which computes both deep causal and strictly causal representations, introduces additional computational overhead, despite sharing parameters and key/value projections and being parallelizable.
To mitigate this overhead in deep models, \model architecture is configured to use a portion of early layers as two-stream attention  rather than all layers, which is studied empirically in the following section.

\begin{proof}
\textit{\textbf{The Composition of Causal and Strictly Causal Layers:}}
We want to show that composing a strictly causal layer with any number of causal ones yields a strictly causal network.
We first prove that the composition of a strictly causal layer and a causal layer results in a network that is, overall, strictly causal. A \emph{strictly causal layer}, $f^{sc}$, where the output at position $n$ depends only on inputs before $n$. Let $g = f^{sc}(x)$, so the output at position $n$ is $g_n = f^{sc}(x_{<n})$. A \emph{causal layer}, $f^{c}$, where the output at position $n$ depends on inputs up to and including $n$. Let $y = f^{c}(g)$, so the output at position $n$ is $y_n = f^{c}(g_{\le n})$. The composite function is $y = f^{c} \circ f^{sc} = f^{c}(f^{sc}(x))$. To prove that this composition is strictly causal, we must show that the final output $y_n$ depends only on the original inputs $x$ at positions less than $n$ (i.e., $x_{<n}$).

The output of the causal layer at position $n$, $y_n$, is a function of inputs $\{g_1, g_2, \dots, g_n\}$. 
Each of these intermediate outputs, $g_i$, is an output from the strictly causal layer $f^{sc}$: 
\begin{itemize}
    \item $g_1$ depends on $x_{<1}$ (an empty set of inputs).
    \item $g_2$ depends on $x_{<2}$ (i.e., just $x_1$)
    \item ....
    \item $g_n$ depends on $x_{<n}$ (i.e., the set $\{x_1, x_2, \dots, x_{n-1}\}$).
\end{itemize}
To find the total dependency of $y_n$, we need to find the set of all original inputs $x$ that are required to compute the set $\{g_1, g_2, \dots, g_n\}$. This is the union of the dependencies of each individual $g_i$:
$$
\scriptsize \text{Dependency}(y_n) = \text{Dependency}(g_1)~\cup~ \text{Dependency}(g_2)~\cup~ \dots ~\cup~ \text{Dependency}(g_n)= \{x_{<1}\} \cup \{x_{<2}\} \cup \dots \cup \{x_{<n}\}
$$
The union of this collection of sets is simply the largest set in the collection, which is $\{x_{<n}\}$.
Therefore, the final output $y_n$ depends only on the inputs strictly preceding it ($x_{<n}$). Hence, the composite function $f^{c} \circ f^{sc}$ is strictly causal.
Also, by induction, the composition of a strictly causal layer with any number of causal ones yields a strictly causal network.
\end{proof}

\begin{figure}[t]
    \centering
    \begin{minipage}[t]{0.99\textwidth}
    \centering
        \includegraphics[trim= 0 25 0 25,clip, angle=0,origin=c, width=0.95\linewidth]{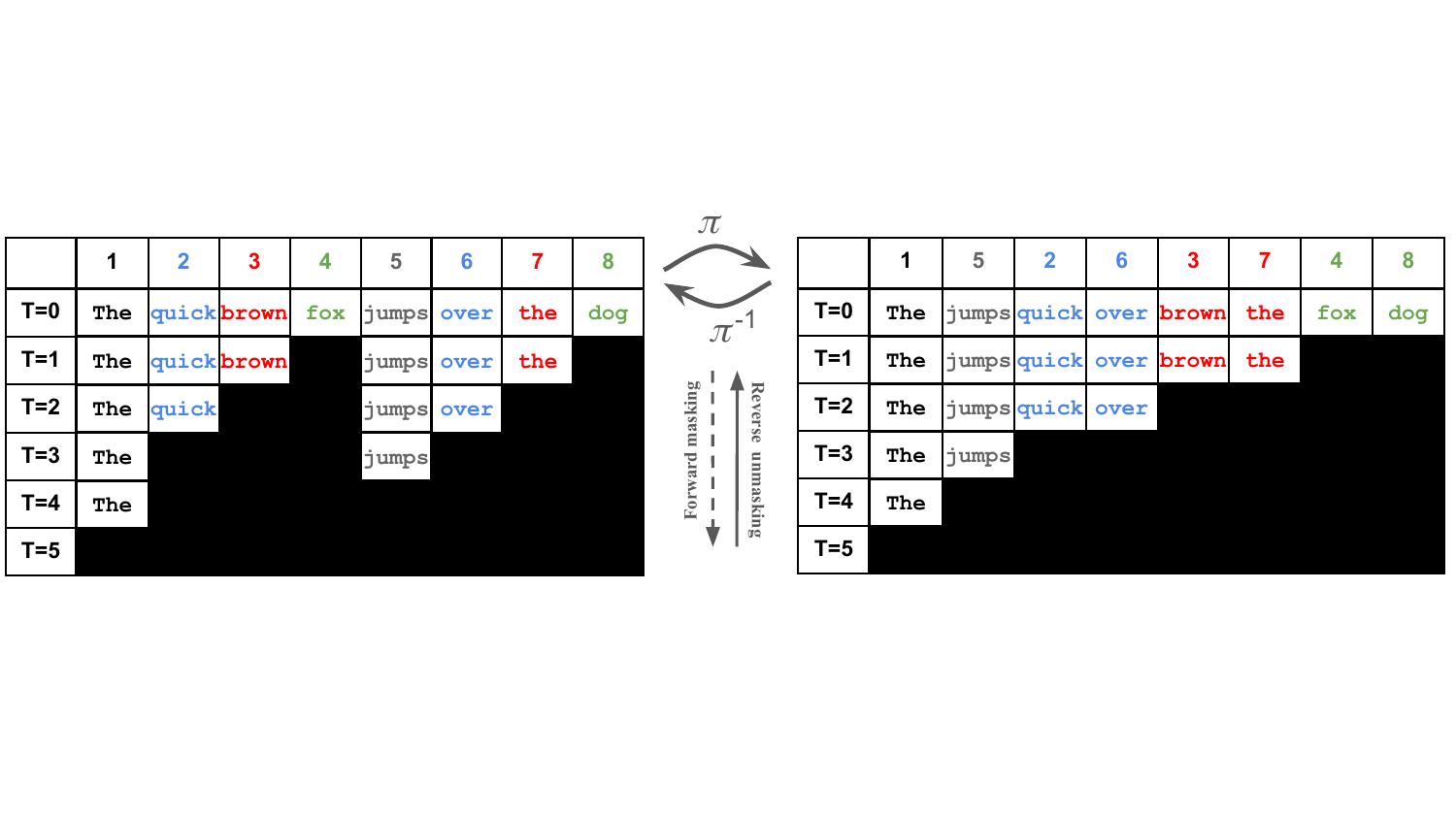}
    \end{minipage}%
    \\
    \vskip -35pt
    \begin{minipage}[t]{0.49\textwidth}
        \centering
        \includegraphics[width=0.8\linewidth]{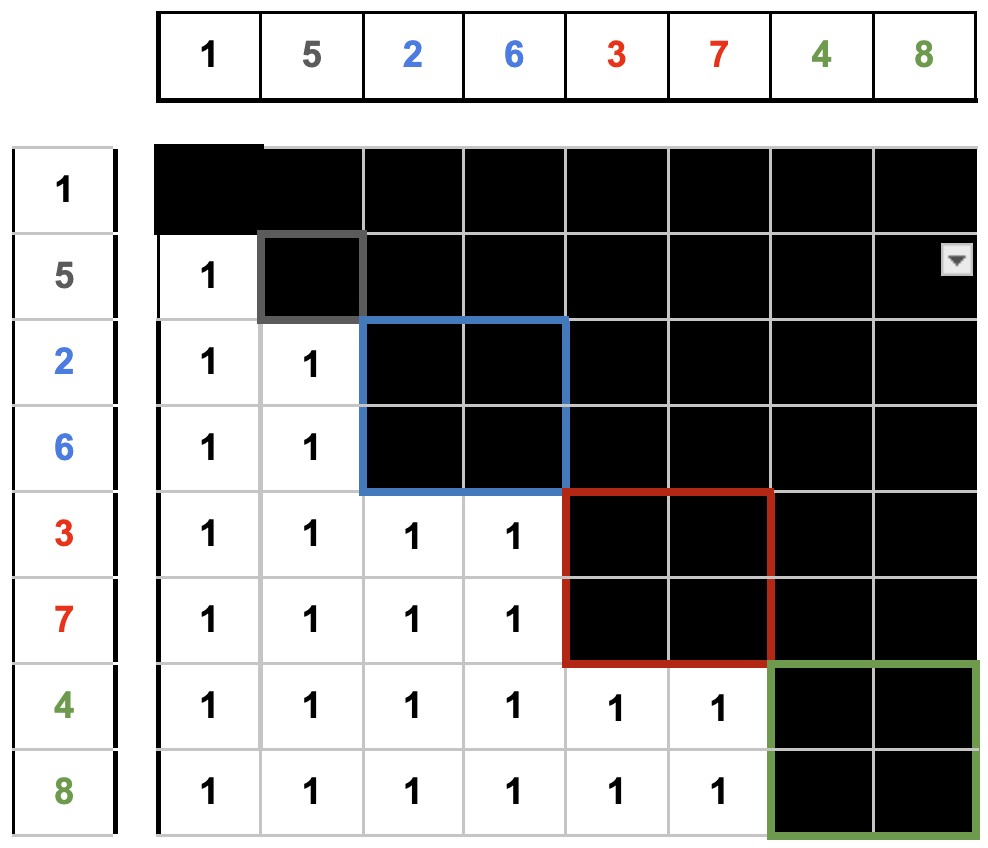}
    \end{minipage}%
    \begin{minipage}[t]{0.49\textwidth}
        \centering
        \setlength{\tabcolsep}{1.1pt} %
        \renewcommand{\arraystretch}{2.}
        \centering
        \vskip -120pt
        \resizebox{1.02\linewidth}{!}{
        \begin{tabular}{llccccccc}
        \footnotesize
            1.~ & \texttt{The} & \_ & \_ & \_ & \_ & \_ & \_ & \_ \\
            2. & \texttt{The} & \_ & \_ & \_ & \texttt{jumps} & \_ & \_ & \_ \\
            3. & \texttt{The} & {\color{blue} \texttt{quick}} & \_ & \_ & \texttt{jumps} & {\color{blue} \texttt{over}} & \_ & \_ \\
            4. & \texttt{The} & {\color{blue} \texttt{quick}} & {\color{red} \texttt{brown}} & \_ & \texttt{jumps} & {\color{blue} \texttt{over}} & {\color{red} \texttt{the}} & \_ \\
            5. & \texttt{The} & {\color{blue} \texttt{quick}} & {\color{red} \texttt{brown}} & {\color{green!60!black} \texttt{fox}} & \texttt{jumps} & {\color{blue} \texttt{over}} & {\color{red} \texttt{the}} & {\color{green!60!black} \texttt{dog}} \\
        \end{tabular}
        }
    \end{minipage}%
    \captionof{figure}{ \footnotesize \label{fig:masking_strided_gen}
    Illustration of the \textit{strided parallel generation}, its diffusion process and masking.
    \textit{(Top Left)}: An instance of masked diffusion process.  
    \textit{(Top Right)}: The causal patterns by permuting the sequence $\vx_{1:N} = \pi(\vz_{1:N})$. 
    \textit{(Bottom left)}: The resulting strictly causal attention mask. 
    \textit{(Bottom right)}: The \textit{strided parallel generation} where steps 1-2 represent the sequential generation of the stream heads, and steps 3-5 show the parallel generation of subsequent tokens. Here, same-colored tokens are generated simultaneously.
    }
\end{figure}

\subsection{Complexity and Latency Analysis} 
\label{apdx:complexity}

In this section, we analyze the theoretical computational cost and empirical latency of the \model{} architecture.

\paragraph{Theoretical Complexity Analysis.} 
The \model{} architecture consists of two distinct components: 
1) A \textit{Two-Stream} attention mechanism for the first $L^{2s}$ layers, which enables the computation of both a Causal stream ($\mX$) and a Strictly Causal stream ($\mG$) to guarantee strict causality and permutation equivariance.
2) A \textit{Standard Causal Stack} for the remaining $L - L^{2s}$ layers. 
Since the attention weights ($\mW^q, \mW^k, \mW^v, \mW^o$) are shared between streams, the total parameter count remains nearly identical to that of a standard model. We compare the Floating Point Operations (FLOPs) of an \model{} layer against a standard AR transformer layer. Let $N$ be the sequence length, $d$ be the hidden dimension, and assume a \method{SwiGLU} Feed-Forward Network (FFN) with an expansion factor of $8/3$ (widely used in modern LLMs like Llama-3).

\begin{itemize}
    \item \textbf{Standard Causal Layer (Single Stream):}
    The Attention Projections ($\mQ, \mK, \mV, \mO$) require $8 N d^2$ with $4 N^2 d$ for the attention mechanism, and the FFN (\method{SwiGLU}) needs $16 N d^2$ FLOPs.
    This brings the total cost per layer to:
    $$ C_{std} = 24 N d^2 + 4 N^2 d $$
    
    \item \textbf{Two-Stream Layer:} 
    This layer computes two hidden states ($\mX$ and $\mG$). While sharing Key/Value states avoids redundant computation, the strictly causal stream $\mG$ requires computing its own specific query ($\hat{\mQ}$) and output ($\mO$) representations ($4 N d^2$), an independent FFN ($16 N d^2$), and the attention scores ($4 N^2 d$). 
    This results in the following additional FLOPs for the Strictly Causal stream:
    $$ C_{extra} = 20 N d^2 + 4 N^2 d $$
    
    \textit{Note:} For the final layer of the Two-Stream stack, we only need to update $\mG$. The computation saved by skipping the update for $\mX$ is approximately equivalent to the additional computation required for the \textit{Prefix Aggregation} layer ($4 N d^2 + 4 N^2 d$).
\end{itemize}

For our default configuration, where the Two-Stream mechanism is applied to the first half of the model ($L^{2s} = L/2$), the total FLOPs ratio relative to a standard AR model is:
$$
    \text{Ratio} = \frac{\frac{L}{2}(C_{std} + C_{extra}) + \frac{L}{2}C_{std}}{L \cdot C_{std}}
$$
Consequently, the architectural overhead is approximately $42\%$ for Dense Operations ($d^2$ term) and $50\%$ for Attention Operations ($N^2$ term).

\paragraph{Empirical Latency.}
We benchmarked the wall-clock training throughput (sequences/second) for a 12-layer model on 4$\times$ NVIDIA H100 GPUs with a sequence length of 1024. Table~\ref{tab:latency} summarizes the results.

\begin{table}[h]
    \centering
    \caption{Training throughput comparison. \model{} incurs a moderate overhead compared to standard AR models, which scales with the number of Two-Stream layers ($L^{2s}$).}
    \label{tab:latency}
    \vspace{0.2cm}
    \begin{tabular}{l c c}
        \toprule
        \textbf{Model Configuration} & \textbf{Throughput (seq/sec)} & \textbf{Relative Overhead} \\
        \midrule
        Standard AR & 448.02 & 1.00$\times$ \\
        \model{} ($L^{2s}=3$) & 406.35 & 1.10$\times$ \\
        \model{} ($L^{2s}=6$) \textit{(Default)} & 368.53 & \textbf{1.22$\times$} \\
        \model{} ($L^{2s}=9$) & 338.89 & 1.32$\times$ \\
        \model{} ($L^{2s}=12$) & 312.98 & 1.43$\times$ \\
        \bottomrule
    \end{tabular}
\end{table}

For our default configuration ($L^{2s}=6$), the training latency is $1.22\times$ that of a standard AR model. However, this overhead is negligible compared to the $3\times$--$8\times$ reduction in total training steps required for convergence relative to baseline masked diffusion models, resulting in a significantly lower \textit{total training FLOPs} footprint. Furthermore, notably, the two streams within a block operate independently and could be sharded via model parallelism to further close this latency gap; however, the results reported here were measured without such optimizations.

\section{Extended Experimental Results, Details, and Ablations} \label{apdx:exp_details}
\subsection{OpenWebText dataset.} \label{sec:OpenWebText}
We trained \model, on the the OpenWebText dataset~\citep{Gokaslan2019OpenWeb} in two sizes—small and medium—matching the parameter counts of \method{GPT-2 ~small} (125M) and \method{medium} (345M)~\citep{radford2019language}: small models use $L = 12$ layers with a hidden dimension of $d=768$ and $12$ attention heads while medium size models are composed of $L = 24$ layers with a hidden size of $d=1024$ and $16$ attention heads. 
The \model architecture is composed of a stack of $L^{2s}$ two-stream layers, followed by $L - L^{2s}$ standard causal layers, as illustrated in \autoref{fig:deep_arch}. 
For our primary models, we configure the first half of the transformer layers as the two-stream module, {i.e.} $L^{2s} = L/2$. 
To study the impact of this choice, we also trained ablation models with different configurations, such as a small model with all layers as two-stream $L^{2s} = 12$ and a medium model with only 1/4 layers as two-stream stack $L^{2s} = 6$.
For the prefix aggregation layer (see \eqref{eq:PA_matrix}), we use sinusoidal positional embeddings~\citep{vaswani2017attention} projected through a two-layer MLP with a latent dimension of d/4. Additionally, all Transformer layers incorporate Rotary Positional Embeddings (RoPE)~\citep{su2024roformer} for relative position awareness.

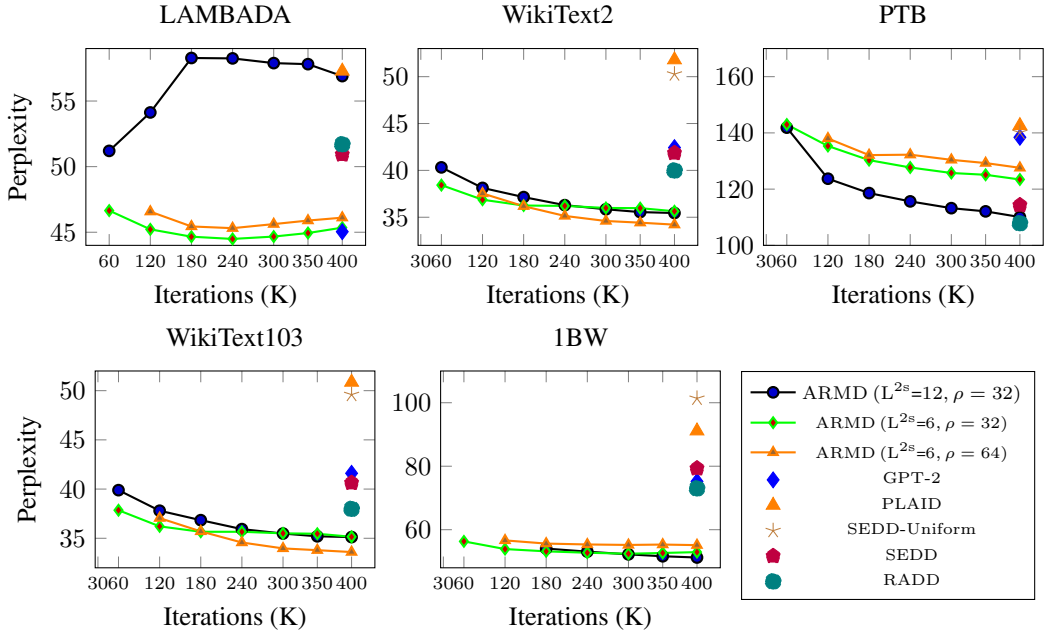
\begin{figure}[htbp]
\centering

\newcommand{\gwidth}{5.3cm}
\newcommand{\gheight}{4.2cm}

\def\plotmodel#1#2#3{%
    \addplot+[mark=#2, color=#3, thick, color=#3] coordinates {#1};%
}

\def\plotother#1#2#3{%
    \addplot+[only marks, mark=#2, color=#3, mark options={fill=#3, mark size=3}] coordinates {(400, #1)};%
}

\begin{minipage}[t]{1.0\textwidth}
\centering
\begin{tikzpicture}
\begin{axis}[
    title={LAMBADA},
    xlabel={Iterations (K)},
    ylabel={Perplexity},
    ymin=44, ymax=59,
    xtick={60, 120, 180, 240, 300, 350, 400},
    xticklabel style={font=\scriptsize},
    width=\gwidth, height=\gheight
]
\plotmodel{
(	60	,	51.20	)
(	120	,	54.13	)
(	180	,	58.28	)
(	240	,	58.25	)
(	300	,	57.89	)
(	350	,	57.81	)
(	400	,	56.90	)
}{*}{black} %
\plotmodel{
(	60	,	46.65	)
(	120	,	45.22	)
(	180	,	44.66	)
(	240	,	44.49	)
(	300	,	44.67	)
(	350	,	44.94	)
(	400	,	45.35	)
}{diamond*}{green} %
\plotmodel{
(	120	,	46.57	)
(	180	,	45.44	)
(	240	,	45.31	)
(	300	,	45.62	)
(	350	,	45.89	)
(	400	,	46.11	)
}{triangle*}{orange} %
\plotother{45.04}{diamond*}{blue}       %
\plotother{57.28}{triangle*}{orange}    %
\plotother{65.40}{star}{brown}          %
\plotother{50.92}{pentagon*}{purple}    %
\plotother{51.70}{oplus*}{teal}         %
\end{axis}
\end{tikzpicture}
\begin{tikzpicture}
\begin{axis}[
    title={WikiText2},
    xlabel={Iterations (K)},
    ymin=32, ymax=53,
    xtick={30, 60, 120, 180, 240, 300, 350, 400},
    xticklabel style={font=\scriptsize},
    width=\gwidth, height=\gheight
]
\plotmodel{
(	60	,	40.32	)
(	120	,	38.13	)
(	180	,	37.15	)
(	240	,	36.29	)
(	300	,	35.85	)
(	350	,	35.55	)
(	400	,	35.45	)
}{*}{black} %
\plotmodel{
(	60	,	38.43	)
(	120	,	36.87	)
(	180	,	36.25	)
(	240	,	36.21	)
(	300	,	35.99	)
(	350	,	35.97	)
(	400	,	35.64	)
}{diamond*}{green} %
\plotmodel{
(	120	,	37.54	)
(	180	,	36.18	)
(	240	,	35.14	)
(	300	,	34.60	)
(	350	,	34.42	)
(	400	,	34.21	)
}{triangle*}{orange} %
\plotother{42.43}{diamond*}{blue}
\plotother{51.80}{triangle*}{orange}
\plotother{50.27}{star}{brown}
\plotother{41.84}{pentagon*}{purple}
\plotother{39.98}{oplus*}{teal}
\end{axis}
\end{tikzpicture}
\begin{tikzpicture}
\begin{axis}[
    title={PTB},
    xlabel={Iterations (K)},
    ymin=100, ymax=170,
    xtick={30, 60, 120, 180, 240, 300, 350, 400},
    xticklabel style={font=\scriptsize},
    width=\gwidth, height=\gheight
]
\plotmodel{
(	60	,	141.87	)
(	120	,	123.71	)
(	180	,	118.59	)
(	240	,	115.64	)
(	300	,	113.23	)
(	350	,	112.09	)
(	400	,	110.07	)
}{*}{black} %
\plotmodel{
(	60	,	142.98	)
(	120	,	135.34	)
(	180	,	130.31	)
(	240	,	127.70	)
(	300	,	125.74	)
(	350	,	125.10	)
(	400	,	123.43	)
}{diamond*}{green} %
\plotmodel{
(	120	,	138.02	)
(	180	,	132.10	)
(	240	,	132.25	)
(	300	,	130.42	)
(	350	,	129.27	)
(	400	,	127.58	)
}{triangle*}{orange} %
\plotother{138.43}{diamond*}{blue}
\plotother{142.60}{triangle*}{orange}
\plotother{140.12}{star}{brown}
\plotother{114.24}{pentagon*}{purple}
\plotother{107.85}{oplus*}{teal}
\end{axis}
\end{tikzpicture}
\end{minipage}
  \hfill
\begin{minipage}[t]{1.0\textwidth}
\centering
\begin{tikzpicture}
\begin{axis}[
    title={WikiText103},
    xlabel={Iterations (K)},
    ylabel={Perplexity},
    ymin=32, ymax=52,
    xtick={30, 60, 120, 180, 240, 300, 350, 400},
    xticklabel style={font=\scriptsize},
    width=\gwidth, height=\gheight
]
\plotmodel{
(	60	,	39.89	)
(	120	,	37.80	)
(	180	,	36.84	)
(	240	,	35.93	)
(	300	,	35.48	)
(	350	,	35.20	)
(	400	,	35.12	)
}{*}{black} %
\plotmodel{
(	60	,	37.85	)
(	120	,	36.21	)
(	180	,	35.66	)
(	240	,	35.66	)
(	300	,	35.50	)
(	350	,	35.46	)
(	400	,	35.15	)
}{diamond*}{green} %
\plotmodel{
(	120	,	37.05	)
(	180	,	35.70	)
(	240	,	34.57	)
(	300	,	33.97	)
(	350	,	33.79	)
(	400	,	33.60	)
}{triangle*}{orange} %
\plotother{41.60}{diamond*}{blue}
\plotother{50.86}{triangle*}{orange}
\plotother{49.60}{star}{brown}
\plotother{40.62}{pentagon*}{purple}
\plotother{37.98}{oplus*}{teal}
\end{axis}
\end{tikzpicture}
\begin{tikzpicture}
\begin{axis}[
    title={1BW},
    xlabel={Iterations (K)},
    ymin=48, ymax=110,
    xtick={30, 60, 120, 180, 240, 300, 350, 400},
    xticklabel style={font=\scriptsize},
    width=\gwidth, height=\gheight,
    legend style=
    {
    font=\tiny, at={(1.6, 1.)}, 
    anchor=north, 
    legend columns=1
    }
]
\plotmodel{
(	180	,	54.05	)
(	240	,	53.09	)
(	300	,	52.23	)
(	350	,	51.63	)
(	400	,	51.24	)
}{*}{black} %
\plotmodel{
(	60	,	56.32	)
(	120	,	53.90	)
(	180	,	53.18	)
(	240	,	52.74	)
(	300	,	52.43	)
(	350	,	52.67	)
(	400	,	52.94	)
}{diamond*}{green} %
\plotmodel{
(	120	,	56.63	)
(	180	,	55.63	)
(	240	,	55.35	)
(	300	,	55.20	)
(	350	,	55.32	)
(	400	,	55.14	)
}{triangle*}{orange} %
\plotother{75.20}{diamond*}{blue}
\plotother{91.12}{triangle*}{orange}
\plotother{101.37}{star}{brown}
\plotother{79.29}{pentagon*}{purple}
\plotother{72.99}{oplus*}{teal}
\legend{
\scriptsize
\method{\model~(L^{2s}\text{=}12, \rho=32)},
\method{\model~(L^{2s}\text{=}6, \rho=32)},
\method{\model~(L^{2s}\text{=}6, \rho=64)},
\method{GPT-2}, \method{PLAID}, \method{SEDD-Uniform}, \method{SEDD}, \method{RADD}
}
\end{axis}
\end{tikzpicture}
\end{minipage}
\caption{Perplexity vs. training steps for \textit{small-size} models.
\label{fig:zero_shot_small}
}
\end{figure}
\begin{figure}[htbp]
\centering

\newcommand{\gwidth}{5.3cm}
\newcommand{\gheight}{4.2cm}
\def\plotmodelm#1#2#3{%
    \addplot+[mark=#2, color=#3, thick, color=#3] coordinates {#1};%
}

\def\plototherm#1#2#3{%
    \addplot+[only marks, mark=#2, color=#3, mark options={fill=#3, mark size=3}] coordinates {(400, #1)};%
}

\begin{minipage}[t]{1.0\textwidth}
\centering
\begin{tikzpicture}
\begin{axis}[
    title={LAMBADA},
    xlabel={Iterations (K)},
    ylabel={Perplexity},
    ymin=34, ymax=52,
    xtick={30, 60, 120, 180, 240, 300, 350, 400},
    xticklabel style={font=\scriptsize},
    width=\gwidth, height=\gheight
]
\plotmodelm{
(	60	,	47.32	)
(	120	,	44.77	)
(	180	,	41.24	)
(	240	,	42.84	)
(	300	,	43.78	)
(	320	,	42.36	)
}{*}{black} %
\plotmodelm{
(	60	,	43.18	)
(	120	,	40.62	)
(	180	,	39.60	)
(	240	,	39.31	)
(	300	,	39.08	)
(	350	,	39.44	)
}{diamond*}{green} %
\plototherm{35.66}{diamond*}{blue}       %
\plototherm{42.77}{pentagon*}{purple}    %
\plototherm{44.10}{oplus*}{teal}         %
\end{axis}
\end{tikzpicture}
\begin{tikzpicture}
\begin{axis}[
    title={WikiText2},
    xlabel={Iterations (K)},
    ymin=25, ymax=34,
    xtick={30, 60, 120, 180, 240, 300, 350, 400},
    xticklabel style={font=\scriptsize},
    width=\gwidth, height=\gheight
]
\plotmodelm{
(	60	,	30.49	)
(	120	,	28.16	)
(	180	,	27.40	)
(	240	,	27.83	)
(	300	,	27.20	)
(	320	,	26.96	)
}{*}{black} %
\plotmodelm{
(	60	,	30.14	)
(	120	,	27.57	)
(	180	,	26.64	)
(	240	,	26.21	)
(	300	,	26.06	)
(	350	,	26.01	)
}{diamond*}{green} %
\plototherm{31.80}{diamond*}{blue}       %
\plototherm{31.04}{pentagon*}{purple}    %
\plototherm{30.60}{oplus*}{teal}         %
\end{axis}
\end{tikzpicture}
\begin{tikzpicture}
\begin{axis}[
    title={PTB},
    xlabel={Iterations (K)},
    ymin=70, ymax=150,
    xtick={30, 60, 120, 180, 240, 300, 350, 400},
    xticklabel style={font=\scriptsize},
    width=\gwidth, height=\gheight
]
\plotmodelm{
(	60	,	122.29	)
(	120	,	110.19	)
(	180	,	108.90	)
(	240	,	105.51	)
(	300	,	103.98	)
(	320	,	104.02	)
}{*}{black} %
\plotmodelm{
(	60	,	122.71	)
(	120	,	105.09	)
(	180	,	100.55	)
(	240	,	99.83	)
(	300	,	97.75	)
(	350	,	97.60	)
}{diamond*}{green} %
\plototherm{123.14}{diamond*}{blue}      %
\plototherm{87.12}{pentagon*}{purple}    %
\plototherm{82.08}{oplus*}{teal}         %
\end{axis}
\end{tikzpicture}
\end{minipage}
  \hfill
\begin{minipage}[t]{1.0\textwidth}
\centering
\begin{tikzpicture}
\begin{axis}[
    title={WikiText103},
    xlabel={Iterations (K)},
    ylabel={Perplexity},
    ymin=24, ymax=34,
    xtick={30, 60, 120, 180, 240, 300, 350, 400},
    xticklabel style={font=\scriptsize},
    width=\gwidth, height=\gheight
]
\plotmodelm{
(	60	,	30.01	)
(	120	,	27.61	)
(	180	,	26.91	)
(	240	,	27.33	)
(	300	,	26.67	)
(	320	,	26.45	)
}{*}{black} %
\plotmodelm{
(	60	,	29.72	)
(	120	,	27.09	)
(	180	,	26.13	)
(	240	,	25.73	)
(	300	,	25.55	)
(	350	,	25.53	)
}{diamond*}{green} %
\plototherm{31.39}{diamond*}{blue}       %
\plototherm{29.98}{pentagon*}{purple}    %
\plototherm{29.29}{oplus*}{teal}         %
\end{axis}
\end{tikzpicture}
\begin{tikzpicture}
\begin{axis}[
    title={1BW},
    xlabel={Iterations (K)},
    ymin=42, ymax=63,
    xtick={30, 60, 120, 180, 240, 300, 350, 400},
    xticklabel style={font=\scriptsize},
    width=\gwidth, height=\gheight,
    legend style=
    {
    font=\scriptsize, at={(1.6, 1.)}, 
    anchor=north, 
    legend columns=1
    }
]
\plotmodelm{
(	60	,	47.07	)
(	120	,	45.44	)
(	180	,	44.62	)
(	240	,	43.60	)
(	300	,	43.54	)
(	320	,	43.42	)
}{*}{black} %
\plotmodelm{
(	60	,	48.16	)
(	120	,	45.49	)
(	180	,	44.57	)
(	240	,	44.28	)
(	300	,	43.91	)
(	350	,	43.78	)
}{diamond*}{green} %
\plototherm{55.72}{diamond*}{blue}       %
\plototherm{61.19}{pentagon*}{purple}    %
\plototherm{60.32}{oplus*}{teal}         %
\legend{
\scriptsize
\method{\model~(L^{2s}\text{=}12, \rho=32)},
\method{\model~(L^{2s}\text{=}6, \rho=32)},
\method{GPT-2},\method{SEDD}, \method{RADD}
}
\end{axis}
\end{tikzpicture}
\end{minipage}
\caption{Perplexity vs. training steps for \textit{medium-size} models.
\label{fig:zero_shot_medium}
}
\end{figure}
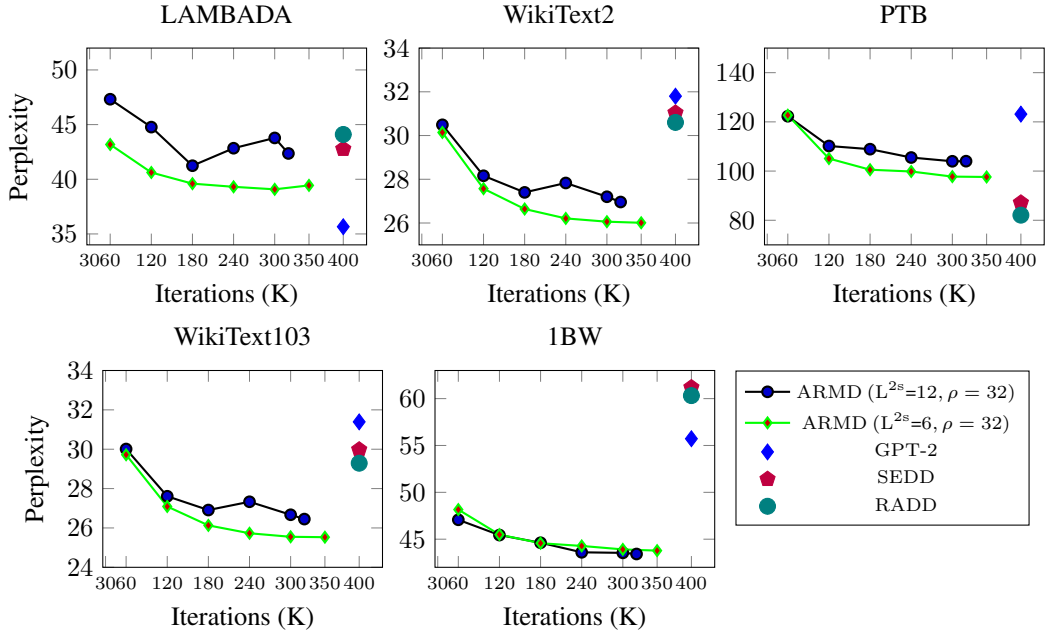

\paragraph{Time-Independent Generative Network.}
Recent studies have indicated that explicit time conditioning in discrete diffusion generative models is either unnecessary or has minimal impact on performance \citep{ou2024your,zheng2024masked,sahoo2024simple}. Consequently, we parameterize our \model model with a time-independent neural network, such that 
$p_\theta(\vx_n \mid [\vx_i \mid  {\mathcal{B}(i) < \mathcal{B}(i)} ]) = \operatorname{NN}^{sc}\left( [\vx_i \mid  {\mathcal{B}(i) < \mathcal{B}(n)} ]\right) [n]$.
We trained \model with equal-length masking blocks of size 1, which corresponds to $T = 1024$ diffusion steps during training, and  set reweighting factor in the objective \eqref{eq:loss_diff_AR} to constant value, $\gamma_t = 1/T$.

We follow the \textbf{training setup} in~\cite{lou2023discrete}, using the OpenWebText dataset~\citep{Gokaslan2019OpenWeb}. All models are trained for 400K iterations with a batch size of 512 sequences, each 1024 tokens long, amounting to approximately 0.5 million tokens per batch. Tokenization is performed using the \texttt{GPT2} tokenizer.

For \textbf{optimizer}, we adopted the AdamW optimizer~\citep{loshchilov2017AdamW} with a learning rate of $3 \times 10^{-4}$ and default hyperparameters ($\beta_1=0.9$, $\beta_2=0.999$, and $\epsilon=1 \times 10^{-8}$), and linear warmup over the first $2000$ iterations. We apply gradient clipping with a norm threshold of $1$, a weight decay of $0.03$, and a dropout rate of $0.02$.
All models were trained on 8$\times$ (or 4$\times$) NVIDIA H100 GPUs. Training \model-small for 100K iterations took approximately 19 hours, while \model-medium required around 50 hours for 100K iterations. For inference and sampling, we used NVIDIA V100 GPUs.

We used a \textit{progressive permutation schedule} during training, where the model is initially trained exclusively on forward-ordered sequences for the first $i_{AR}$ iterations. After this phase, we gradually introduce partial permutations by randomly selecting a small subset of tokens—starting with a single token—and permuting their positions within the sequence. The number of permuted tokens is progressively increased up to a maximum of $\rho$ by iteration $i_{perm}$.
This curriculum enables the model to become increasingly robust to token permutations while still retaining the left-to-right order as the primary structural prior. Since natural language inherently exhibits a strong left-to-right structure, leveraging this inductive bias as guidance ensures that the model retains strong performance in capturing forward dependencies.

We compare \model against several established and recent generative models with similar parameter counts, including:  \method{GPT-2} \citep{radford2019language}, \method{D3PM} \citep{austin2021structured}, \method{PLAID} \citep{gulrajani2023likelihood}, \method{SEDD} \citep{lou2023discrete} and \method{RADD} \citep{ou2024your}. 
To evaluate language modeling performance of \method{\model}, we report zero-shot perplexity on widely-used benchmarks:
LAMBADA~\citep{paperno2016lambada}, Penn Tree Bank (PTB)~\citep{Marcus1993BuildingAL}, WikiText2, WikiText103,~\citep{merity2016pointer},  and One Billion Words (1BW)~\citep{chelba2014billion}.

As  the results in \autoref{tbl:zero_shot_small_med} and \autoref{fig:zero_shot_small}, \autoref{fig:zero_shot_medium} demonstrate, small (medium) \model achieves state-of-the-art performance on four (three) benchmarks, while 
\method{\model-small} outperforms \method{GPT-2~small} across all benchmarks.
Notably, it outperforms diffusion baselines in most benchmarks with significantly less training iterations, highlighting the training efficiency that is a key advantage of our model.

 \begin{table*}[htbp]
  \centering
\footnotesize
\caption{
  \label{tbl:zero_shot_small_med_ablations}
    \footnotesize
    \textbf{Zero-shot language modeling perplexity ($\downarrow$) 
    }:
    \method{D3PM}~\citep{austin2021structured},  \method{PLAID}~\citep{gulrajani2023likelihood} and \method{SEDD-Uniform}~\citep{lou2023discrete} are  discrete diffusion models, while \method{SEDD}~\citep{lou2023discrete} and \method{RADD}~\citep{ou2024your}  are masked language models.
The number of training steps for each model is presented in parenthesis. 
 $^*$Results for baseline diffusion models are taken from the reported upper bounds in ~\cite{lou2023discrete,ou2024your}.
The best result for each dataset is shown in \textbf{bold}, and the second-best is \underline{underlined}.
}
\vspace{-5pt}
\setlength{\tabcolsep}{4pt}
\renewcommand{\arraystretch}{1.2}
  \begin{tabular}{lccccc}
\toprule
\quad  \method{Method} & \method{LAMBADA} & \method{WikiText2} & \method{PTB} & \method{WikiText103} & \method{1BW} \\ 
\midrule
\midrule
\multicolumn{6}{l}{\textsl{small size models}} \\
\quad {\method{GPT-2^*}} & \underline{45.04} & 42.43 & 138.43 & 41.60 & 75.20\\
\arrayrulecolor{black!30}\midrule
\quad {\method{D3PM^*}} & 93.47 & 77.28 & 200.82 & 75.16 & 138.92\\
\quad {\method{PLAID~(600K)^*}} & 57.28 & 51.80 & 142.60 & 50.86 & 91.12\\
\quad {\method{SEDD-Uniform ~(400K)^*}} & 65.40 & 50.27 & 140.12 & 49.60 & 101.37\\ 
\arrayrulecolor{black!30}\midrule
\quad {\method{SEDD ~(400K)^*}}& 50.92 & 41.84& \underline{114.24}& 40.62 & 79.29\\
\quad {\method{RADD ~(400K)^*}} & 51.70& {39.98}& \textbf{107.85}& {37.98}& 72.99\\
\arrayrulecolor{black!30}\midrule
\multicolumn{6}{l}{$\{ {L^{2s}=6,~ \rho=32} \}$}  \\
\quad {\method{\model~(60K)}}	&	46.65	&	38.43	&	142.98	&	37.85	&	56.32	\\
\quad {\method{\model~(180K)}}	&	\textbf{44.66}	&	\underline{36.25}	&	130.31	&	\underline{35.66}	&	\underline{53.18}	\\
\quad {\method{\model~(400K)}}	&	45.35	&	\underline{35.64}	&	123.43	&	\underline{35.15}	&	\underline{52.94}	\\
\multicolumn{6}{l}{$\{ {L^{2s}=6,~ \rho=64} \}$}  \\
\quad {\method{\model~(60K)}}	& 47.16	&	37.83	&	152.75	&	37.34	&	57.26	\\	
\quad {\method{\model~(180K)}}	& 45.44	&	36.18	&	132.10	&	35.70	&	55.63	\\	
\quad {\method{\model~(400K)}}	& 46.11	&	\textbf{34.21}	&	127.58	&	\textbf{33.60}	&	55.14	\\
\multicolumn{6}{l}{$\{ {L^{2s}=12,~ \rho=32} \}$}  \\
\quad {\method{\model~(180K)}}	&	58.28	&	37.15	&	118.59	&	36.84	&	54.05	\\
\quad {\method{\model~(400K)}}	&	56.90	&	35.45	&	110.07	&	35.12	&	\textbf{51.24}	\\
\arrayrulecolor{black!100}\midrule
\midrule
\multicolumn{6}{l}{\textsl{medium size models}} \\
\quad {\method{GPT-2^*}} & \textbf{35.66}& 31.80& 123.14& 31.39& {55.72}\\
\arrayrulecolor{black!30}\midrule
\quad {\method{SEDD ~(400K)^*}} & 42.77 & 31.04 & 87.12 & 29.98 & 61.19\\
\quad {\method{RADD ~(400K)^*}} & 44.10& 30.60& \textbf{82.08}& 29.29& 60.32\\
\arrayrulecolor{black!30}\midrule
\multicolumn{6}{l}{$\{ {L^{2s}=12,~ \rho=32} \}$}  \\
\quad {\method{\model ~(60K)}}	& 43.18	&	30.14	&	122.71	&	29.72	&	48.16	\\
\quad {\method{\model ~(120K)}}	&	40.62	&	\underline{27.57}	&	105.09	&	\underline{27.09}	&	\underline{45.49}	\\
\quad {\method{\model ~(300K)}}	&	\underline{39.08}	&	\textbf{26.06}	&	97.75	&	\textbf{25.55}	&	\underline{43.91} \\
\multicolumn{6}{l}{$\{ {L^{2s}=6,~ \rho=32} \}$}  \\
\quad {\method{\model ~(60K)}}	& 47.32	&	30.49	&	122.29	&	30.01	&	47.07	\\
\quad {\method{\model ~(120K)}}	& 44.77	&	28.16	&	110.19	&	27.61	&	45.44 \\	
\quad {\method{\model ~(300K)}}	& 43.78	&	27.20	&	103.98	&	26.67	&	\textbf{43.54}	\\	
\arrayrulecolor{black!100}\bottomrule
  \end{tabular}
\end{table*}

\subsection{Ablation Studies} \label{sec:ablation_studies}
We conduct ablation studies for language models trained on OpenWebText dataset to evaluate the effects of the two-stream attention layers and the progressive permutation schedule. Specifically, we train \model-small under the following configurations:
\begin{enumerate}[itemsep=0.1em, topsep=0pt]
    \item Small \model with $L^{2s} = 12$: all layers use two-stream attention, while the remaining layers use standard causal attention.
    \item Medium \model with $L^{2s} = 6$: a first quarter of layers use two-stream attention.
    \item Permutation schedule with \textit{32 permutes}: The number of permuted tokens gradually increased, starting at iteration $i_{AR}$= $9$K and reaching the maximum $\rho$=$32$  by iteration $i_{perm}$=$48$K.
    \item Permutation schedule with \textit{64 permutes}: $\rho = 32$, with $i_{AR} = 40$K and $i_{perm} = 120$K.
\end{enumerate}

As shown in \autoref{tbl:zero_shot_small_med_ablations}, setting $L^{2s} = L/2$ is enough for achieving strong performance while reducing the number of two-stream layers to $L^{2s} = L/4$ slightly compromises performance—though still comparable and better than the baseline methods—while offering improved computational efficiency. Increasing the number of permutation steps from 32 to 64 further improves performance on some datasets.

\subsection{One Billion Words dataset.} \label{sec:LM1B}
We also train language models on One Billion Words (LM1B) dataset ~\citep{chelba2014billion}.
We follow the training and architecture setting of  \cite{he2022diffusionbert, lou2023discrete, sahoo2024simple}: the models' size match GPT-2 small (12 layers with a hidden dimension of 768 and 12 attention heads),
\texttt{bert-base-uncased} tokenizer is applied.
\model uses two-stream modules for its first half of the layers, \ie $L^{2s} = L/2 = 6$.
For training, the sequences are padded and truncated to a length of 128 with batch size of 512.
Models are trained for 32B tokens which correspond to 1M steps in diffusion.
($i_{AR}=9K$ iterations).
For progressive permutation schedule of \model, starting from iteration $i_{AR}=100K$, we gradually introduce random permutations by increasing the number of shuffled tokens up to a maximum of $\rho=8$ by iteration $i_{perm}=300K$.

For \textbf{optimizer}, similar to training for OpenWebText, we use the AdamW optimizer with a learning rate of $3 \times 10^{-4}$, a weight decay of $0.03$, a linear warmup schedule over the first $2000$ iterations with gradient clipping at $1$. We use a dropout rate of $0.02$.

\subsection{Sample Generation}  \label{sec:sample_gen} %

A recent study by~\citet{zheng2024masked} found that numerical errors in \texttt{float32} precision during Gumbel-based categorical sampling—commonly used in MDMs—can lead to deceptively low sample perplexity while producing low token diversity (i.e., low sample entropy). This effectively mimics the effect of using a lower sampling temperature. 
To mitigate this effect, they suggest using higher precision (\texttt{float64}) during sampling.
Therefore, we employ \texttt{float64} precision for sampling across all MDMs in our experiments. Additionally, we increase the sampling temperature until the generated sequences achieve a level of token entropy comparable to that of MDMs, ensuring diversity in the outputs.
Unigram entropy is computed for each generated sequence by measuring the entropy over its token distribution, and then averaged across all sampled sequences.

The performance of sequential and parallel sample generation of the \model compared to baseline diffusion models are  presented in \autoref{tab:parallel_gen}, 
and some examples of the generated texts are presented in the following section.

\clearpage
\begin{framed}
\footnotesize
\textbf{Sample 1 for model: \model-small and sequential generation ($T=1024$):}

 keys to functionality. The little messages they’re communicating need to be called to send messages, and then they need to have their counterparts in the star system also able to use them. Otherwise, they can’t receive messages and lose their abilities. That solution, on the other hand they needed to do something, too – name it “coming tomorrow.”

When I go through the ancient texts of the Greek philosopher Archytas about sorrows, sorrows of immortality, the Devil and the light of the world, there’s a lot more detail there than you might think. There are echoes in all that subject matter going on. In the past, there were pages long with the same two or three character states. Scholars have pointed out how chirality defines rules when it’s fun to do so. If you really want to be remembered, you just write down something. If you want to be remembered in a particular divided field of study, you have to write down something that is significantly different than the classification you have already or in your mind. Everyone has something different to write down. But having that is equally important: we can’t make the ancient system, could we? Somebody has to explain something in writing, or at least break boundaries, and often that is the task. And that brings us to one of the most important facts of.

There’s an old story. A chariot was moving in confused circles. The chariot stopped actively in the middle of the road and got overhead. It suddenly began to go out in a circle, a par rated Stamina Lightweight Robot. At the time of the random shaking in the par has happened, at least, it wasn’t a metal robot, it is strong, a chance-wise, and it is capable of lifting massive objects. But since the collision the game can of handstand and cardiac division, and the light weight of the case opened another possibility (heat also wants to protect the body from heat). Here’s what happened at the moment of the accident the mechanical object continued to hit the slingshot part of the chariot and the effect was catastrophic? The flaming wreckage of the chariot with its wreckage on the right side blew off the other side of the vehicle. The child who suffered very badly and ended up in a coma went completely unconscious. His organs were all but gone, he had died in a fiery explosion. Something else that happened in the chariot was that the truck that drove into it was damaged, misaligned. Perhaps that went completely unexploded, there are a lot of effects of this kind that can be sustained over little hours. And still. “You saw something or acted (for some unknown reason) or maybe (it was) another thing that happened. Yeah, this wasn’t easy.”

If they were looking for much more polarity, you’d want to imagine these events arising in the wormway of time. The customer of a window might fly in the time, he’s probably heard something used-up about the J4 computer being able to carry out the job of computer hacking. Isn’t that parallelism the major reason why lower level memories teach us about time? An example of self-resemblance to this kind of stuff is the one that Adam used to reach to Kylie from the depths of hell. It was a very rude kind of memory-reconnaissance exercise. I remember seeing a horrified form of air and darkness, and as you’ve seen, when it was brought awake, the joy of seeing other beings would cease. I can only imagine how someone may have been unable to imagine it. You know, there would be additional legs possible, it would be really rich in fun. But for a moment Kylie said: you’re a potential victim of such a meteorological disaster, unless you jumped out of a plane like me.

The story of the moon, being frozen and not yet crushed by the impact, keeping itself up, does such things all the time. There’s a new story happening here, and here’s how it relates to sending a basic need to take care of yourself to you. And this story does include a classic example of how your initiated self can take care of you, er.

Sentient Intelligence

This is a note from the book Electromagnetic Times: The Quest for Ultrafast Time 1000

In an age when, say, a spacecraft needs every power available to send it around the Sun, why not send it now when it’s coming close enough to shut down?

Within that brief chapter, you’ll see how the revelation of a more interstellar solar system with thousands of starships emitting thousands of battle buddies and their radar was all but confirmed by everyday data regarding the planet itself, proving a great assumption. In that context, it is wiser to move on from the question of life in

\end{framed}

\clearpage
\begin{framed}
\footnotesize
\textbf{Sample 1 for model: \model-small and parallel generation $S=2$ ($T=512$):}

 clearance interviews of developers ensuring that devices can easily be deployed, receive updates, and communicate as seamlessly as possible in standard operating system systems – it’s a big, huge deal. There are myriad intersection points, mustering a multitude of requests to support and eventually eliminating the need to build dev tools.

First, a quick look at what distinguishes open source development from proprietary projects

Representative— This includes all aspects of the design, implementation, maintenance, support, and cooperation of open source software.

Project engineer— This includes the technician, technical support, and writing of the full project documentation.

Before we get into the performance of customization thereof, let me point out general reasons why open source software offers a higher level of performance than proprietary IT systems:

Built on a single OS platform

Suffice it to say that we have bend, pluck, claw, weave, and gouge in line with open source solutions today; however, we thankfully get to use a completely different set of OS and programming tools.

Controlling on the server has been a core part of the backbone of Open Source IT for some time; but for many IT departments this requiring platforms like SUSE is often more headed towards the future and smaller teams will be more willing to put development effort in. I’m sorry, Linux has come a long way, and that’s not to say that OSX, MacOS, SporTV, and even Windows couldn’t solve their problems.

Open Source has also come a long way on its own tools. Tools are the foundation of Open Source IT, and many are based on Linux. Linux was PC oriented and easy to get started, but an open source Linux development pipeline will allow developers greater freedom and scope to use a wider range of tool and subsystem concepts.

Open Source also allows developers to refactor, remove, change, improve, or maintain minimum levels of functionality with greater scope; paying developers the loyalty to a single operating system, and not the other way around.

A good example is Open Container Initiative (OCI). Open Container Initiative (OCI) has in the past been a non-profit venture and even distributed, yet for a long time, its developers weren’t affiliated with IT departments.

This gave them an edge in unit driven projects; however, the issues with projects like Open Container Initiative (OCI) lead them to destroy it, since they made the massive mistake of having merged with Open Container Initiative (OCI) in 2009.

Tacoma, Wash.-based Open Container Initiative (OCI) is an open source project that is trying to integrate the concept of containers into systems operating under Linux. Its priorities include:

Enhance the ability of open source containers to efficiently solve application problems

Support a variety of cross-platform, eco-system building workloads

Provide an anti-blue screen of death all over enterprise IT systems

Allow the easy integration of containers into systems operating under Linux

It’s necessary to discuss infrastructure associated with containers:

Compatibility and application architecture: Which operating systems you run should be all the same, is critical for any “single” OS?

Distribution lifecycle: Which OS should we build for? Whether we build for Windows, Linux, or OS X? Does adding few layers of management matter?

Forming an ecosystem from scratch: What open source technology should we pick? Should there be a vendor certification set up?

Other aspects with consideration:

Which platform should we target?

For what use limitation should we place a proprietary solution runtime on or a low tier operating system?

If the goal of Open Source is to provide versatility, with high-performance and low overhead, shouldn’t we not execute code on the same platform?

With a global set of tools and APIs, Open Source can provide solutions to all types of problems and at the same time eliminate the need for many layers/distributions to replace the one (or less) available.

Let’s look at some other generally applicable principles that you identify for Open Source in terms of developer capabilities:

A balanced organization that balances maintaining and improving the infrastructure with support and complexity that advantages technology (or tactics) over the other.

Management of open source software in code can help to resolve congestion and improve quality and deployment of system updates and patches.

An enterprise-wide “transient” mind-set, so that the intent isn’t to hit costs of maintaining an implementation but rather to “just in case” without any risk.

It’s important to note that policies that intuit or enforce how systems are to operate should also be applied individually based on corporate security and compliance considerations. This also extends to the smaller organizations that just don’t care about cost if the benefits outweigh value for the upgrade

\end{framed}

\clearpage
\begin{framed}
\footnotesize
\textbf{Sample 2 for model: \model-small and parallel generation $S=4$ ($T=256$):}

A new ABC musical, "American Legend," which is set in the 19th century British West Indies, features Jonathan Tucker (Andrew Joseph). Tucker, played by local favorite Richard J. Browne, also stars.

Week Two [ edit ]

Week Three marks the return of Alan Scheck (who starred in the 1960 Broadway production of Exchange). The show is two-parter for the first time in Disney's Broadway history. The playing concludes with the Producers meeting in the old Planet Filmy Railroad to discuss the subject of their enterprise.

Week Four [ edit ]

Week Four marks the return of Woody Harrelson, Tim Curry, and Mary J. Blige as Woody. They are joined by a young three-time Academy Award winner in Christopher Lloyd, and a visiting professor from the University of British Columbia in the title role.

Week Five [ edit ]

Week Five marks the return of Duo Aguirre, Benito Villalobos, and Normal Brown Parker as Yancy, turned into Gigul! In 1999, the cast and crew of the Broadway musical's Land of the Millennium sought out possible movie roles. They signed up the actors for a Jack London revival of The House of Waverley "as a" Broadway, Inc. The Broadway musical went on to gross more than \$340 million on Broadway.

Week One [ edit ]

Week One is the latest in a series of productions that focus on the history of Disney characters done that way. One brief mention of Disney characters and the story of the imaginative property comes from The Little Mermaid, the first musical directed and co-produced by Buzz Lightyear and Rodney Dangerfield as an original animated musical. A short skip-step at the Performance Hall (where the musical is set) shows the promotion in effect that Disney cantor William Waterhouse predicted the stage musical would become the year's first nationwide success story. The trio of musicals usually consist of one musical and take place in approximately 700,000 locations. In each such production, character names are listed beginning with Disney characters.[citation needed]

Covers [ edit ]

Although it has premiered in Broadway's original production space, the musical has one rare exception: the production only played in Broadway's first act during the Cedar Fair Fair – January 16, 1993, shortly before the holiday season began.[note 2]

Seasons' premiere [ edit ]

The musical's first two seasons – one musical, the other, the musical (man) are neatly co-presented on Broadway at the ArcLight Bank Theatre. The production opened in Winterthur to positive reviews and a standing ovation. On November 12, 2013, the Broadway production opened at David Letterman's CBS Radio show, Airline. It made its debut at the Disneyland Resort in July 2014. It began its third season on November 7, 2015. It is based on the theme "I'll Wear You," from Disney's film The Lion King. The four seasons of the show were released in DVD format in September and October 2012. It is a high-energy, high character musical,[5] consisting almost entirely of singing and stunts (or "aladdin", depending on the question of aladdin's entertainment value) from 1970 to the present day. The musical was presented by Jack Levitt and Michael Neuman in January 2013. Its production orchestra has had a long presence in the Hollywood Hills since the '60s, and groups such as The Repertory Company of Los Angeles and Holst Music will provide accompaniment. This cast turns out the best-known and richest musical in the history of the Hamptons.

Troll Confidential [ edit ]

Goodman is joined on stage by Tom DeVos and Daranassi of Burbank on September 28. Troll executive producer and regular associated with the production, Kevin Reed, will serve as executive producer on both productions.

Sales for Twitter user \#Aladdin–a Jeff Faison–inspired animated computer comedy–leaked going back to September 4, 2015 highlighting the talents behind the film and its characters.

Season 6 [ edit ]

The Broadway musical’s 10th season is in previews and has been in production since August 17, 2016. This season includes themed episodes and both recent Broadway musicals, like Rent, A24’s Once Upon a Time in America and Casino. While part of the legacy for the Broadway musical of 2010, Beauty and the Beast, its treatment and DVD release, was, at the time, unprecedented in terms of what could be, the show stuck around.

Series screenwriters Jay Kricfalusi and Adam Pally have written six total scripts with proceeds going to the Scouts of America. Kricfalusi continues in his post-New York City book, How to Succeed in Business Without Really Trying.

The musical will open a

\end{framed}

\end{document}